\documentclass{article}

\RequirePackage{fancyhdr}
\RequirePackage{color}
\RequirePackage{algorithm}
\RequirePackage{algorithmic}
\RequirePackage[numbers]{natbib}
\RequirePackage{eso-pic} 
\RequirePackage{forloop}

\usepackage{microtype}
\usepackage{graphicx}
\usepackage{subcaption}
\usepackage{amsmath}
\usepackage{amssymb}
\usepackage{amsfonts} 
\usepackage{booktabs} 
\usepackage{listings}
\usepackage{xcolor}
\definecolor{codegreen}{rgb}{0,0.5,0}
\definecolor{codeblue}{rgb}{0.25,0.5,0.5}
\definecolor{codegray}{rgb}{0.6,0.6,0.6}

\DeclareMathOperator{\Tr}{trace}

\usepackage[nonatbib,preprint]{nips_2018}

\title{Learning Permutations with Sinkhorn Policy Gradient}

\author{
  Patrick Emami \\
	Department of Computer and \\
    Information Science and Engineering\\
    Gainesville, FL, 32611\\
  \texttt{pemami@ufl.edu} \\
  \And
  Sanjay Ranka \\
	Department of Computer and \\
    Information Science and Engineering\\
    Gainesville, FL, 32611\\
  \texttt{sanjayranka@gmail.com} \\  
}

\begin{document}

\maketitle

\begin{abstract}
Many problems at the intersection of combinatorics and computer science require solving for a permutation that optimally matches, ranks, or sorts some data. These problems usually have a task-specific, often non-differentiable objective function that data-driven algorithms can use as a learning signal. In this paper, we propose the Sinkhorn Policy Gradient (SPG) algorithm for learning policies on permutation matrices. The actor-critic neural network architecture we introduce for SPG uniquely decouples representation learning of the state space from the highly-structured action space of permutations with a temperature-controlled \textit{Sinkhorn layer}. The Sinkhorn layer produces continuous relaxations of permutation matrices so that the actor-critic architecture can be trained end-to-end. Our empirical results show that agents trained with SPG can perform competitively on sorting, the Euclidean TSP, and matching tasks. We also observe that SPG is significantly more data efficient at the matching task than the baseline methods, which indicates that SPG is conducive to learning representations that are useful for reasoning about permutations.
\end{abstract}

\section{Introduction}
\label{submission}

Learning to solve combinatorial optimization problems from data has applications in many fields. As a motivating example, consider planar minimum/maximum weight matching. Given a bipartite graph with vertices represented as points in the plane, the objective is to find a permutation matrix that matches vertices such that the sum of Euclidean distances amongst all match pairs is minimized/maximized. Other related combinatorial problems include graph matching \cite{caetano2009learning,nowak2017note}, ranking \cite{cao2007learning,adams2011ranking}, and data association in multi-object tracking \cite{Wojke2017simple,milan2017online,milan2017data}. Data association can be cast as a minimum weight matching problem; given a sequence of raw images containing object detections, the objective is to find the optimal matching between detections in adjacent images to form object tracks.
\par
In this paper, we propose a data-driven algorithm for the task of learning permutations based on a policy gradient method from the reinforcement learning (RL) literature. Data-driven approaches to solving combinatorial problems involve training a model on a dataset of problem instances drawn from a distribution, so that the model is able to score highly with respect to the task-specific objective on a test set of instances drawn from the same distribution. Models trained with supervised learning can perform well by relying on traditional loss functions such as mean squared error \cite{vinyals2015order,vinyals2015pointer}, cross-entropy error \cite{nowak2017note}, or mean squared error augmented with task-specific objectives \cite{milan2017data}. However, obtaining large quantities of solved problem instances to build a labeled dataset is not always feasible.

Recent empirical results indicate that a task-specific objective can be used as the sole learning signal. Most notably, \cite{bello2016neural} uses an encoder and decoder with attention, trained with REINFORCE \cite{williams1992simple}, to solve combinatorial problems that have a sequential nature (e.g., the Euclidean TSP). This was recently extended to take advantage of the graph structure of the inputs by replacing the encoder with a graph attention layer \cite{kool2018attention}. A DQN \cite{mnih2015human}-inspired algorithm proposed in \cite{dai2017learning} also learns graph embeddings of problem instances, for the Euclidean TSP, Maximum Cut, and Minimum Vertex Cover problems. We have developed SPG, a policy gradient method designed for the class of combinatorial problems involving permutations. Indeed, the action space of a policy trained with SPG is the discrete set of $N \times N$ permutation matrices ($\mathcal{P}_N$). In contrast to similar learning-based approaches \cite{bello2016neural,dai2017learning,kool2018attention}, SPG is not restricted to learning policies that emulate a greedy heuristic. We demonstrate that SPG is able to learn to sort integers, produce near-optimal matchings for maximum weight matching (MWM), and find tours of competitive length on the NP-Hard Euclidean Traveling Salesman Problem (TSP). On the MWM task, SPG is more data-efficient and outperforms baseline methods when scaling up to larger problem sizes. 

\subsection{Learning permutations with the Sinkhorn-Knopp algorithm}

Learning permutations is challenging for two main reasons. First, the number of permutations grows factorially in the size of the problem. Second, the non-differentiability of $\mathcal{P}_N$ prevents learning algorithms from directly using backpropagation for end-to-end training. Due to \cite{adams2011ranking}, a truncated version of the Sinkhorn-Knopp algorithm \cite{sinkhorn1964relationship}, which maps a square matrix to a doubly-stochastic matrix (all rows and columns sum to one), was derived for use as a layer in a composable, end-to-end differentiable model. This \textit{Sinkhorn layer} produces continuous and differentiable relaxations of permutation matrices. Recently, \cite{latentPerms2018} built upon this work and introduced Gumbel-Sinkhorn networks for performing inference on probabilistic models that have latent permutation variables. Gumbel-Sinkhorn networks combine Sinkhorn layers with a temperature-controlled softmax gradient estimator for discrete variables \cite{jang2016categorical,maddison2017concrete} to infer permutations from data. We use temperature-controlled Sinkhorn layers to develop a deep neural network architecture for a policy network that can be trained with SPG. We also describe a technique for removing bias induced by the continuous relaxation. The rest of our paper is structured as follows. In the next section, we introduce the requisite background for explaining SPG. In Section 3, we provide details on SPG. In Section 4, we present findings from ablation studies and the main set of experiments. In Section 5, we discuss related works and conclude. 
\section{Background}
\label{sec:background}
\paragraph{Notation.} We use capital letters to represent scalars (e.g., $N$, $K$), bolded capital letters for matrices (e.g., $\mathbf{M}$), script capital letters for sets (e.g., $\mathcal{S}$), lowercase letters for elements of a set (e.g., $s \in \mathcal{S}$), and bolded lowercase letters for vectors (e.g., $\mathbf{b}$). Occasionally it will be necessary to introduce a concept using notation from another paper; we will point out these instances to maintain clarity. 

\subsection{Problem setting}
The learning framework we most closely follow is the average reward contextual bandits setting. Formally, at each step, there is a context which consists of a state $s \in \mathcal{S}$ and a set of actions $\mathcal{A}$ available to an agent. Here, $\mathcal{A} = \mathcal{P}_N$. States are instances of combinatorial problems of size $N$ drawn from a distribution $\rho$.  As an example, consider the task of sorting ten numbers; $\mathcal{S}$ is the set of all 10! orderings of the ten numbers, $\rho$ is a uniform distribution over $\mathcal{S}$, and $\mathcal{A}$ is the set of all $10 \times 10$ permutation matrices. Unlike the standard contextual bandits setting, the number of ``bandit arms'' scales factorially in the problem size $N$ and we are interested in finding a deterministic policy $\pi: \mathcal{S} \rightarrow \mathcal{A}$ that does not take the entire context as input. It might be possible to also formulate this problem with a stochastic policy $\pi(a \vert s)$ using, e.g., the tractable, differentiable density over permutations based on the reparameterization trick proposed recently in \cite{linderman2017reparameterizing}. The recently introduced Gumbel-Sinkhorn distribution does not have a tractable density \cite{latentPerms2018}, and hence we leave this line of inquiry for future work.
  
The immediate reward, or the return, that the agent receives from the environment is $r(s,\pi(s))$. If we parameterize the policy with parameters $\theta$, the agent's objective can be defined as finding $\theta^*$ that maximizes the average reward $J(\pi_\theta) = \mathop{\mathbb{E}}_{s \sim \rho} [r(s, \pi_\theta(s))]$. Note that the objective can also be written in terms of the \textit{regret}, or the expected amount of return the agent ``missed out on`` by selecting sub-optimal actions.

\subsection{Deterministic policy gradient methods}

We now show how the off-policy actor-critic deterministic policy gradient algorithm \cite{silver2014deterministic}, originally derived for RL, can be formulated for our contextual bandits setting. Let the deterministic target policy be $\pi_\theta$. The action-value function $Q(s,a)$ induced by policy $\pi_{\theta}$, defined as the return for choosing action $a$ in state $s$ under deterministic policy $\pi_{\theta}$, is equal to $r(s, \pi_{\theta}(s))$. In Section 4.2 of \cite{silver2014deterministic}, the off-policy deterministic policy gradient is defined for an arbitrary behavior policy $\beta$ (e.g., an $\epsilon$-greedy policy for encouraging exploration) as  

\begin{align}
\label{eq:DPG}
\nabla_\theta J_\beta (\pi_\theta) \approx \mathop{\mathbb{E}}_{s \sim \rho} \big[ \nabla_\theta \pi_\theta (s) \nabla_a Q(s,a) \big \vert_{a = \pi_\theta(s)} \big],
\end{align}
where $Q(s,a)$ is estimated by a differentiable critic $Q_{\theta'}(s,a)$ in practice with parameters $\theta'$, fit with off-policy experience from $\beta$. Note that throughout this paper, we assume that the regularity conditions for the Markov Decision Process in \cite{silver2014deterministic} also hold for our contextual bandits setting. To update the critic parameters $\theta'$, we can descend the gradient of the following mean squared error loss:
\begin{equation}
\label{eq:critic-loss}
\nabla_{\theta'} \mathcal{L}_{\theta'} = \mathop{\mathbb{E}}_{s \sim \rho} \big[ \big (r(s, \beta(s) ) - Q_{\theta'}(s, \beta(s)) \big) \nabla_{\theta'} Q_{\theta'}(s, \beta(s)) \big ].
\end{equation}

Equations \ref{eq:DPG} and \ref{eq:critic-loss} can be derived by setting the RL episode horizon $T$ in \cite{silver2014deterministic} to one, i.e., we are equating our contextual bandits setting to 1-step RL. The extension of our proposed policy gradient method to episodic RL is possible and is left for future work. In practice, the policy and action-value functions in Equations \ref{eq:DPG} and \ref{eq:critic-loss} are implemented as deep neural networks, as is the case with the actor-critic paradigm used in the deep deterministic policy gradient (DDPG) algorithm \cite{lillicrap2015continuous}. As a result of using nonlinear function approximation for learning the actor and critic, the policy gradient (Equation \ref{eq:DPG}) is no longer exact and bias is introduced. The critic is not updated with a 1-step temporal difference rule as it is in DDPG; hence, target networks are not needed to stabilize learning. In Section \ref{sec:SPG}, we will provide complete algorithmic details on how we make use of Equations \ref{eq:DPG} and \ref{eq:critic-loss} to implement SPG. In the next section, we introduce the technique used to relax $\mathcal{P}_N$ to a continuous and differentiable set, which is necessary for computing the policy gradient.

\subsection{Temperature-controlled Sinkhorn layer}
To use the deterministic policy gradient defined in the previous section, we need a continuous and differentiable approximation of $\mathcal{P}_N$. Recently, \cite{latentPerms2018} proposed to combine Sinkhorn layers \cite{adams2011ranking} with a temperature-controlled continuous relaxation inspired by the Gumbel-Softmax trick \cite{jang2016categorical,maddison2017concrete}. We first describe Sinkhorn layers for positive square matrices $\mathbf{X}$, which we hereafter refer to as the Sinkhorn operator $S^L(\cdot)$. Using the notation of \cite{latentPerms2018}, the Sinkhorn operator is defined recursively for $1 \leq i \leq L$ as
\begin{align}
S^{0} &= \exp(\mathbf{X}) \\
S^{i} &= \mathcal{T}_c (\mathcal T_r (S^{i-1}(\mathbf{X}))),
\end{align}
where $\mathcal{T}_c^{j,k}(\mathbf{X}) = \frac{\mathbf{X}_{j,k}}{\sum_l \mathbf{X}_{l,j}}$ is the column normalization operator and $\mathcal{T}_r^{j,k}(\mathbf{X}) = \frac{\mathbf{X}_{j,k}}{\sum_l \mathbf{X}_{j,l}}$ is the row normalization operator. The gradients with respect to the input can be computed efficiently by unrolling each row and column normalization \cite{adams2011ranking}. One of the main results from \cite{latentPerms2018} is that, by introducing a temperature parameter $\tau$,  applying $S^L(\cdot/\tau)$ to $\mathbf{X}$ produces the doubly-stochastic matrix $\mathbf{M}$ whose entries tend to zero or one in the limit as $\tau$ approaches zero. In practice, $L$ and $\tau$ are treated as hyperparameters; we discuss how they are selected for our experiments in Section \ref{sec:experiments-ablation} and Appendix \ref{sec:app-hyper}. Example code for a stable implementation of $S^L(\cdot/\tau)$ in log-scale to mitigate numerical errors is provided in Appendix \ref{sec:app-SPG-details}. In the next section, we describe $S^{L}(\cdot/\tau)$ is used within the SPG actor network.

\section{Sinkhorn Policy Gradient}
\label{sec:SPG}

\begin{figure}
\centering
\includegraphics[scale=0.125]{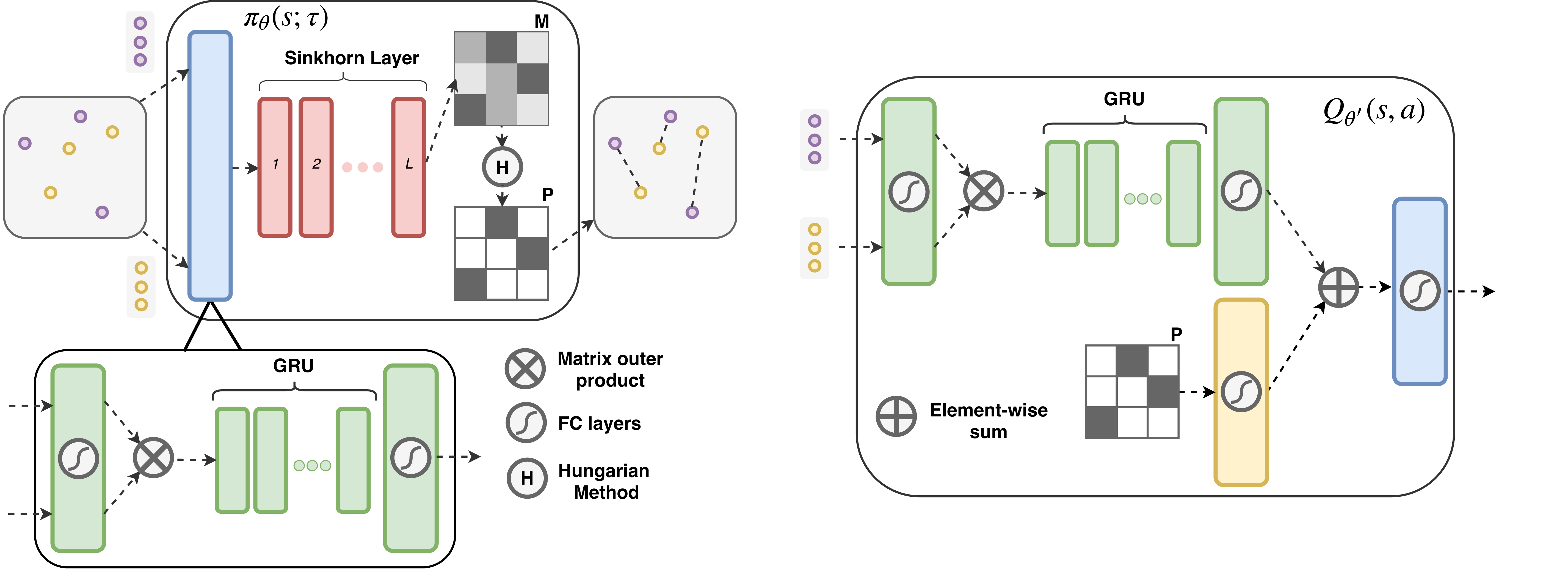}
\caption{Visual summarization of SPG+Matching's actor (left) and critic (right) network architectures; full details are provided in Appendix \ref{sec:app-SPG-details}. In the actor network, two sets of objects are processed by a shared embedding layer, after which the matrix outer product of the two embeddings is computed. The fused embedding is fed into a GRU for representation learning over all possible match pairs. The output is passed to a Sinkhorn layer, which produces the doubly-stochastic matrix $\mathbf{M}$. The Hungarian algorithm is used to round $\mathbf{M}$ to a permutation matrix $\mathbf{P}$. The gradient of the actor network is taken with respect to $\mathbf{M}$, ``bypassing" $\mathbf{P}$. The critic network uses the same embedding layer for the state, and the action is embedded with a fully-connected layer. These embeddings are fused with an element-wise sum, and mapped via a linear layer to a scalar. Best viewed in color.\label{fig:SPG-arch}}
\end{figure}

In this section, we describe SPG, an off-policy deterministic policy gradient algorithm for the action space of $\mathcal{P}_N$. We also introduce a novel actor-critic neural network architecture for SPG, as well as a technique for reducing the policy gradient bias induced by the continuous relaxation. 

We relax $\mathcal{P}_N$ to the continuous set of $N \times N$ doubly-stochastic matrices with $S^{L}(\cdot/\tau)$. The actor network takes the state as input, typically an $N \times K$ matrix when $\mathcal{S} = \mathbb{R}^{N \times K}$, embeds it into a high-dimensional space, and then uses $S^{L}(\cdot/\tau)$ to output a doubly-stochastic matrix $\mathbf{M}$. Note that since $\tau$ is fixed, the entries of $\mathbf{M}$ are not binary. At the start of training, the entries of $\mathbf{M}$ will be close to $1/N$, and as the actor gets more ``confident" in its action selection, $\mathbf{M}$'s values get pushed closer to either zero or one. We round $\mathbf{M}$ to the nearest permutation matrix $\mathbf{P} = \text{H}(\mathbf{M})$ with the $O(n^3)$ Hungarian algorithm \cite{NAV:NAV3800020109,munkres1957algorithms}. Here, ``nearest'' technically means the permutation $\mathbf{P}$ that maximizes $\Tr (\mathbf{P}^{\intercal}\mathbf{M})$. 

Given $\mathbf{P}$ and $\mathbf{M}$, we can now redefine the policy and critic loss gradients by slightly adjusting Equations \ref{eq:DPG} and \ref{eq:critic-loss}. The new policy gradient looks like

\begin{align}
\vspace{-1em}
\label{eq:SPG-actor-gradient-2}
\nabla_\theta J_\beta (\pi_\theta) \approx \mathop{\mathbb{E}}_{s \sim \rho} \big[ \nabla_\theta \pi_\theta (s) \nabla_{a} Q_{\theta'}(s,a) \big \vert_{a = \mathbf{M}} \big]
\end{align}
and the new critic loss gradient is
\begin{equation}
\begin{aligned}
\label{eq:SPG-critic-gradient-2}
\nabla_{\theta'} \mathcal{L}_{\theta'} = \mathop{\mathbb{E}}_{s \sim \rho} \big[ \big (r(s, \mathbf{P} ) - Q_{\theta'}(s, \mathbf{P})\big) \nabla_{\theta'} Q_{\theta'}(s, \mathbf{P}) \big ].
\end{aligned}
\end{equation}

In Equation \ref{eq:SPG-actor-gradient-2}, we need to use the continuous relaxation $\mathbf{M}$ for computing $\nabla_\theta \pi_\theta (s)$ because $\mathbf{P}$ is not differentiable with respect to $\theta$. $\mathbf{M}$ is also used to compute the critic action-gradient because $\nabla_{a} Q_{\theta'}(s,a) \vert_{a = \mathbf{P}}$ is zero almost everywhere, and furthermore $Q_{\theta'}(s,\mathbf{P})$ has discontinuities. This heuristic of ``bypassing" $\mathbf{P}$, i.e., treating it like an identity function when computing the policy gradient, is inspired by the straight-through gradient estimator \cite{bengio2013estimating}. The complete training loop for SPG, which resembles the one used by DDPG, is provided in Algorithm \ref{alg:SPG} in Appendix \ref{sec:app-alg}. Like DDPG, SPG uses a replay buffer when computing the gradients in the backwards pass. We next describe how we can remove most of the bias that the relaxation introduces into the policy gradient.

\begin{figure*}
\centering
\begin{subfigure}{0.25\textwidth}
\centering
\includegraphics[width=\columnwidth]{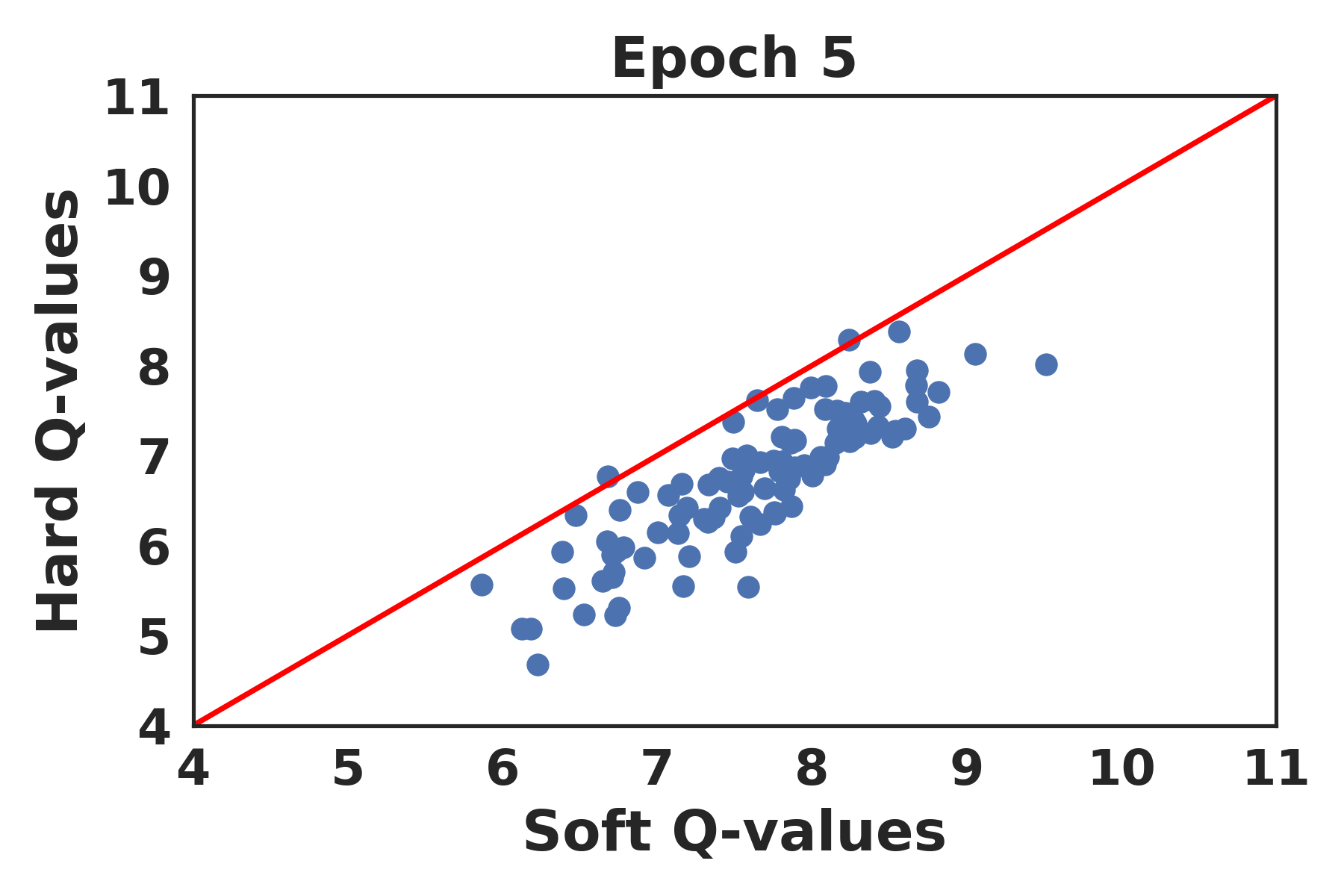}
\caption{\label{fig:biased-pg-epoch-5}}
\end{subfigure}%
\begin{subfigure}{0.25\textwidth}
\centering
\includegraphics[width=\columnwidth]{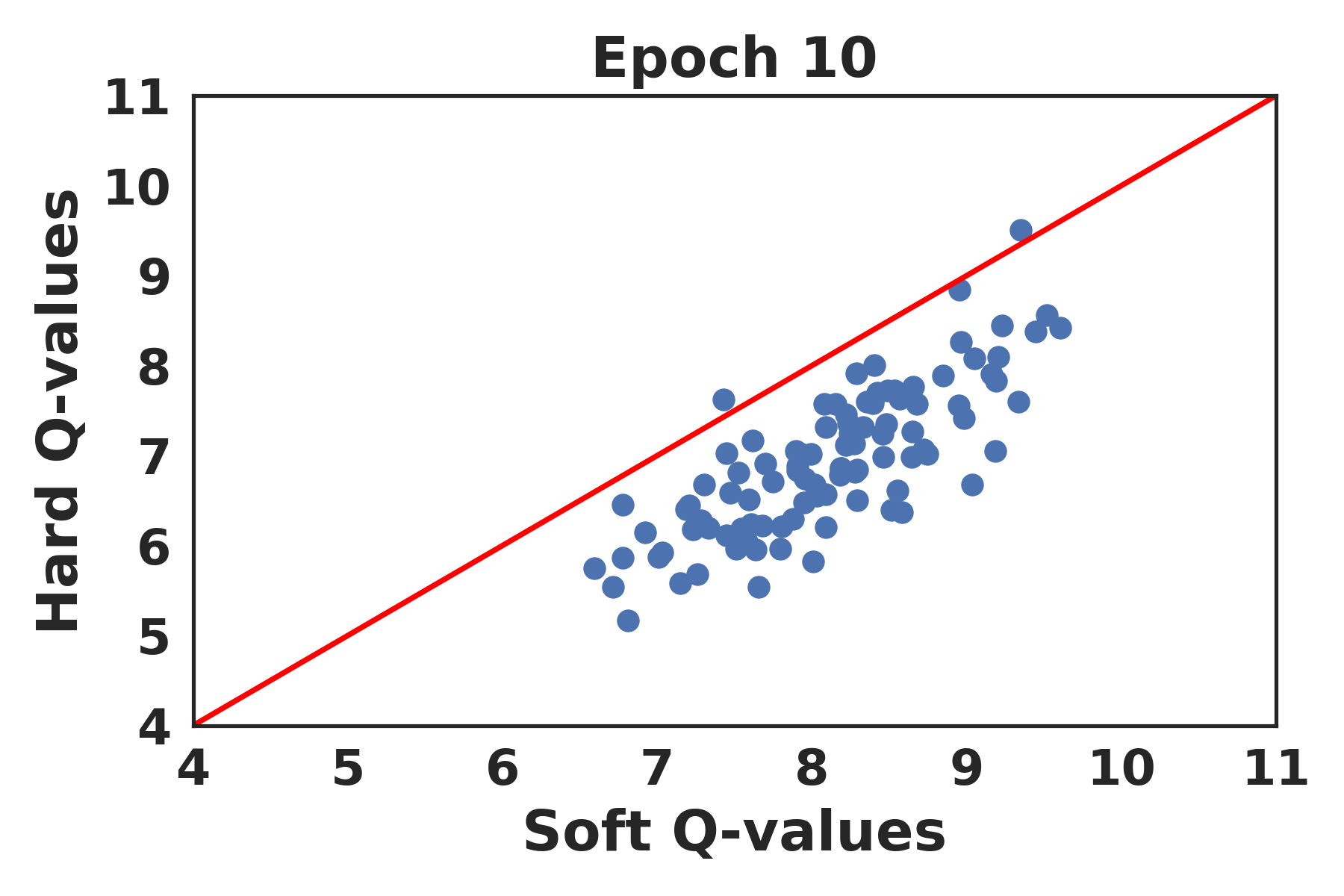}
\caption{\label{fig:biased-pg-epoch-10}}
\end{subfigure}%
\begin{subfigure}{0.25\textwidth}
\centering
\includegraphics[width=\columnwidth]{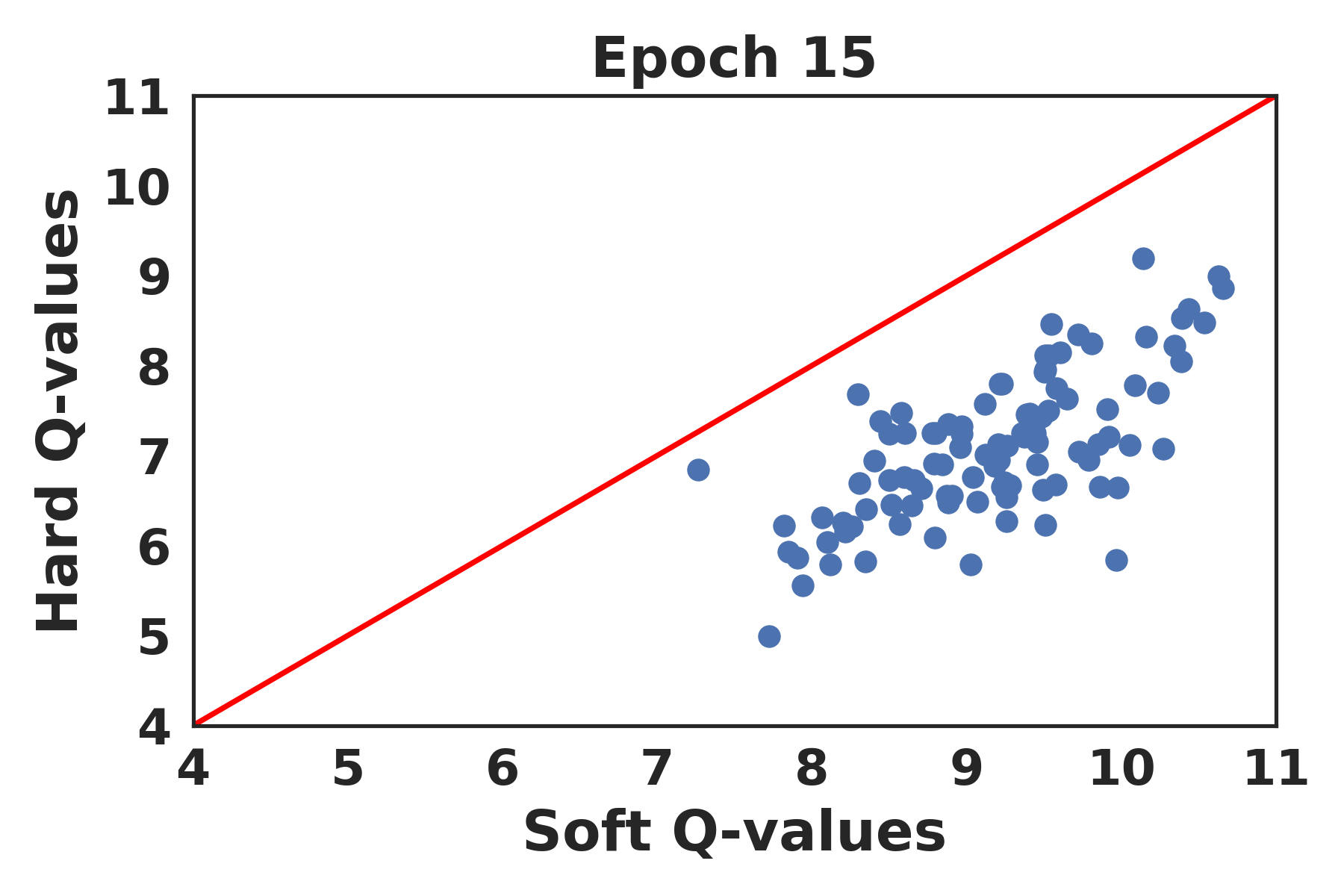}
\caption{\label{fig:biased-pg-epoch-15}}
\end{subfigure}%
\begin{subfigure}{0.25\textwidth}
\centering
\includegraphics[width=\columnwidth]{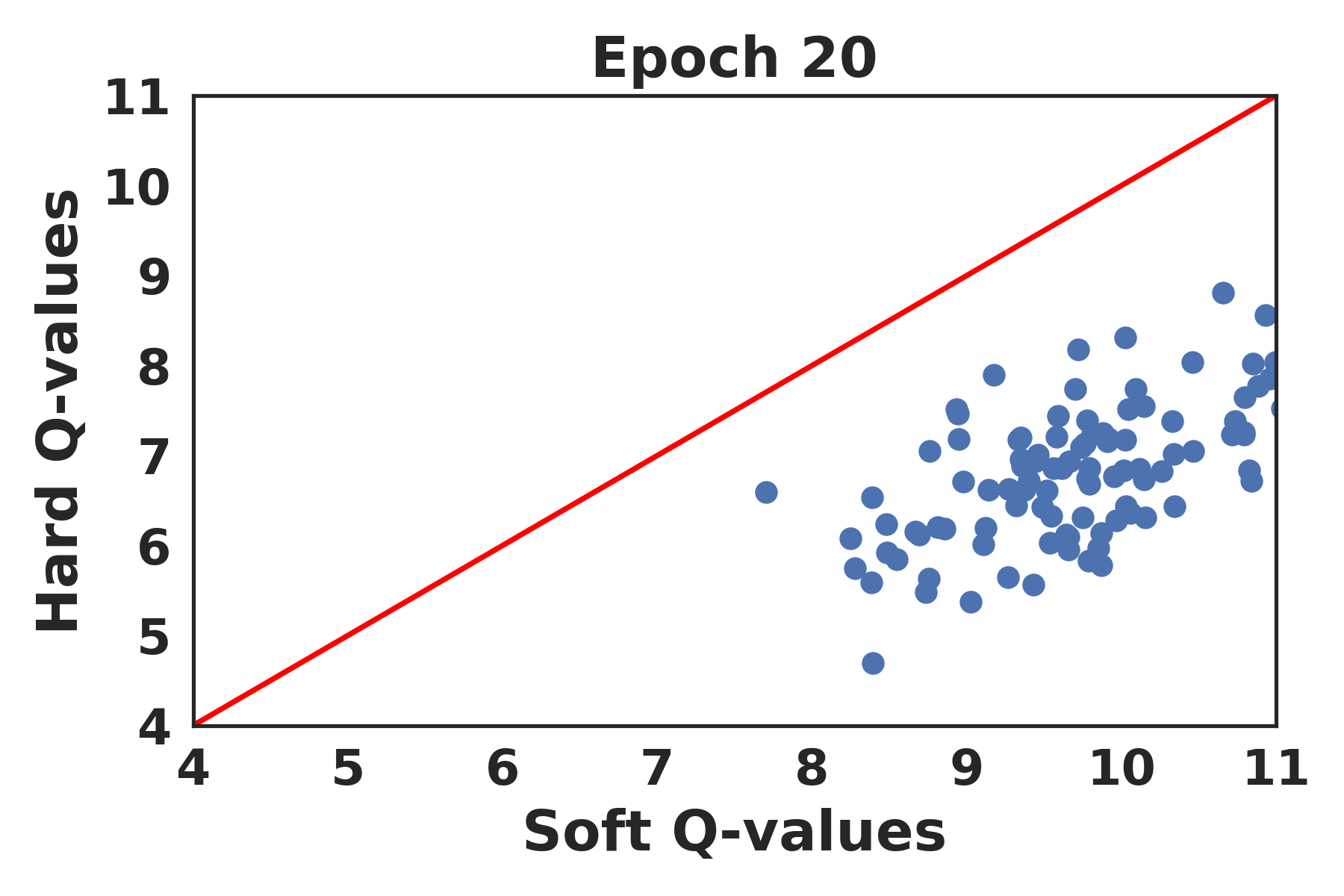}
\caption{\label{fig:biased-pg-epoch-20}}
\end{subfigure}%
\\
\begin{subfigure}{0.25\textwidth}
\centering
\includegraphics[width=\columnwidth]{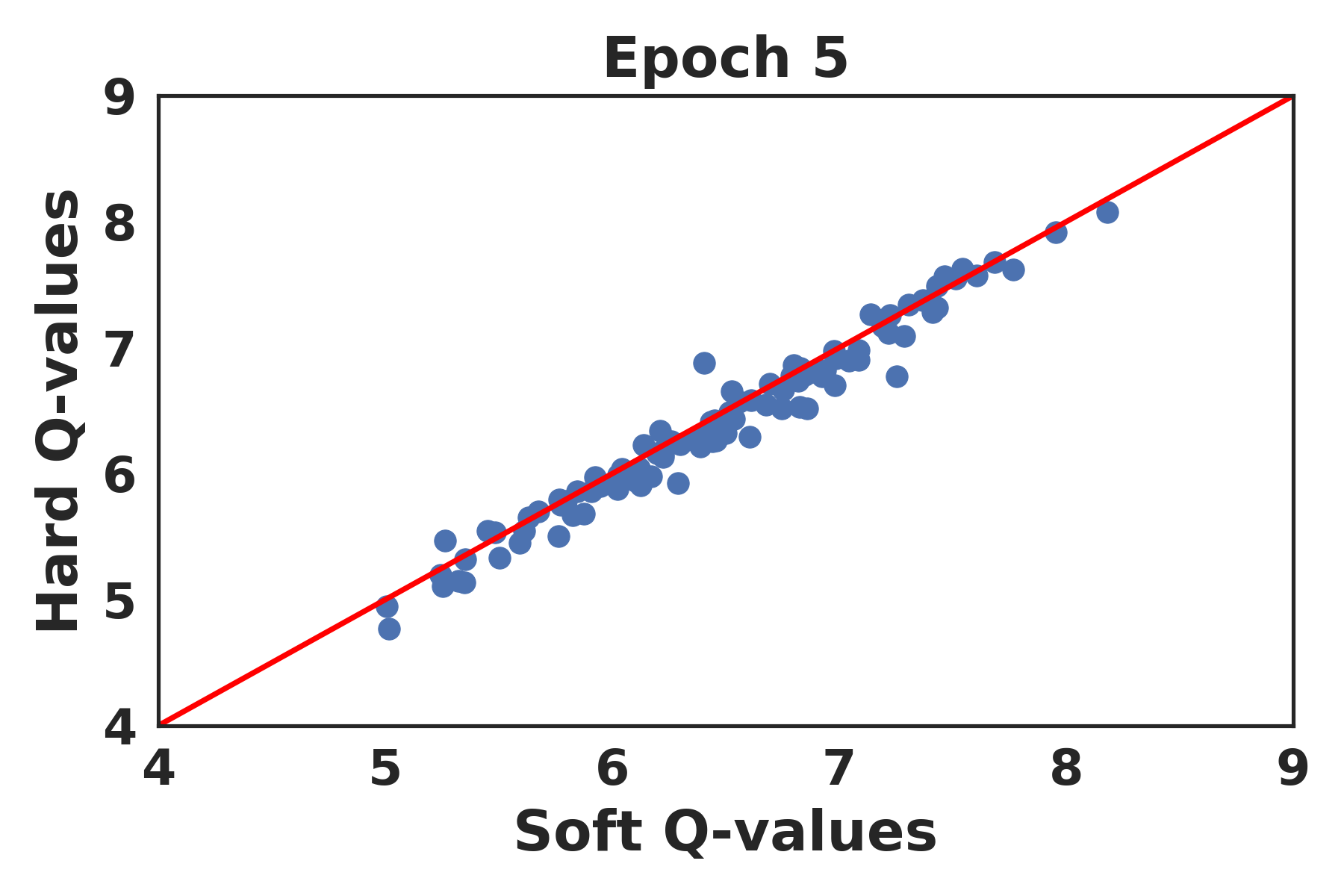}
\caption{\label{fig:unbiased-pg-epoch-5}}
\end{subfigure}%
\begin{subfigure}{0.25\textwidth}
\centering
\includegraphics[width=\columnwidth]{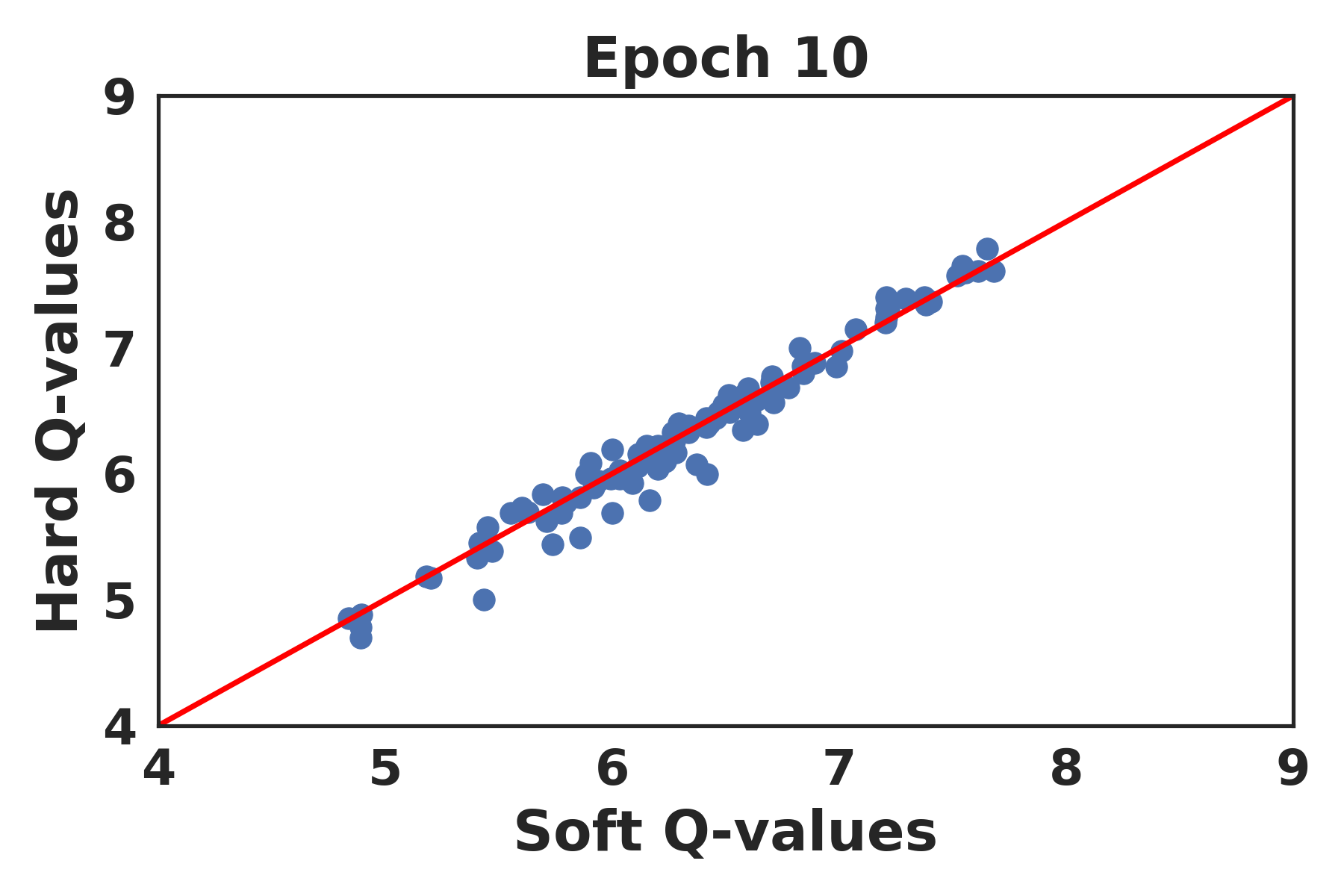}
\caption{\label{fig:unbiased-pg-epoch-10}}
\end{subfigure}%
\begin{subfigure}{0.25\textwidth}
\centering
\includegraphics[width=\columnwidth]{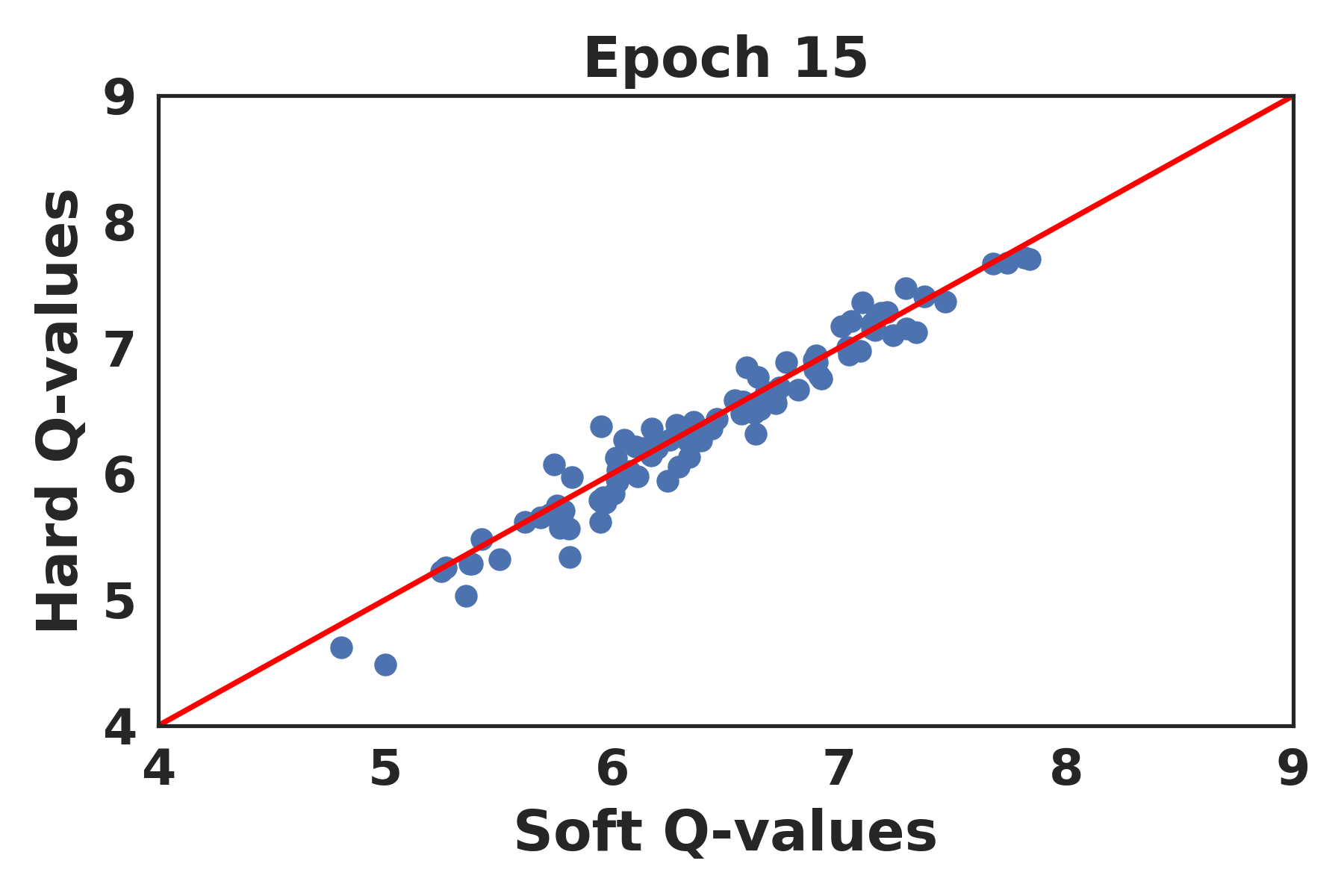}
\caption{\label{fig:unbiased-pg-epoch-15}}
\end{subfigure}%
\begin{subfigure}{0.25\textwidth}
\centering
\includegraphics[width=\columnwidth]{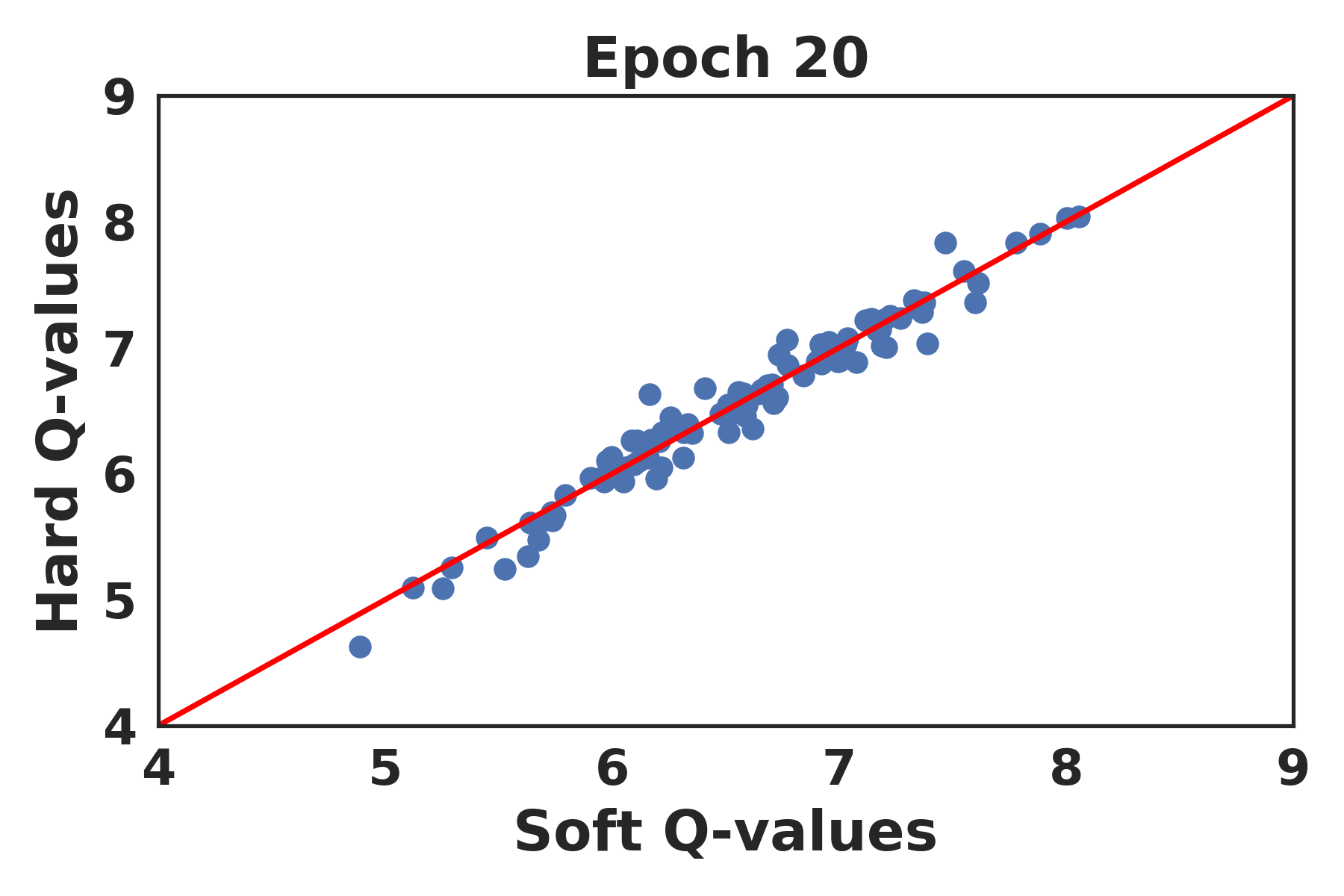}
\caption{\label{fig:unbiased-pg-epoch-20}}
\end{subfigure}%
\caption{(a-d) The top row compares predicted Q-values for ``hard" (discrete) and ``soft" (relaxed) permutations from a critic trained \textit{without} the additional penalty term on MWM-10. Each column shows Q-values from a set of 100 random states after [5,10,15,20] epochs of training (left to right). The diagonal line visualizes $y=x$. (e-h) The bottom row compares predicted Q-values for a critic trained \textit{with} the additional penalty term. The linear relationship between the ``hard" and ``soft" Q-values shows the penalty term's effectiveness.\label{fig:penalty-Q-values}}
\end{figure*}

\paragraph{De-biasing the policy gradient.}In our initial experiments, we observed that the actor network's improvement in terms of average reward would stagnate quite early during training; however, the critic loss would continue to decrease. Notice that the policy gradient is defined using the critic action-gradient $\nabla_{a} Q(s,a) \vert_{a = \mathbf{M}}$ taken with respect to \textit{continuous actions} $\mathbf{M}$, and the critic loss is defined for \textit{discrete actions} $\mathbf{P}$. Since the Q-values for the continuous actions were not being updated by the critic loss, the critic action-gradient was not accurately approximating the direction of maximum reward improvement (see Figure \ref{fig:penalty-Q-values}a-d). To de-bias the policy gradient, we added an auxiliary term to the critic loss that treats the problem of approximating the Q-values for the discrete actions with the Q-values for the continuous actions as a regression. The critic loss with the penalty for dissimilarity between Q-values is 

\begin{equation}
\label{eq:SPG-critic-gradient-3}
\mathcal{L}_{\theta'} = \text{MSE} \big (r(s, \mathbf{P}), Q_{\theta'}(s, \mathbf{P}) \big) + \text{MSE} \big( \texttt{stop\_grad}\big( Q_{\theta'}(s, \mathbf{P}) \big) , Q_{\theta'}(s, \mathbf{M}) \big),
\end{equation}

where MSE is short-hand for the mean squared error. Equations \ref{eq:SPG-actor-gradient-2} and \ref{eq:SPG-critic-gradient-3} are the key components of the SPG algorithm. We provide a geometric interpretation as to why the penalty term helps the critic action-gradient correctly approximate the direction of maximum reward improvement in Appendix \ref{sec:app-critic-penalty}. In Section \ref{sec:critic-penalty-ablation}, we show the penalty term's effectiveness with an ablation study. 
\paragraph{Actor-critic architecture.} SPG uses deep neural networks to implement the actor $\pi_\theta(s;\tau)$ and the critic $Q_{\theta'}(s,a)$. The SPG+Matching architecture is shown in Figure \ref{fig:SPG-arch}. For non-matching combinatorial problems where the only input is a single set of objects, the architecture differs in that there is only a single input to the embedding layer, the GRU is bidirectional, and the matrix outer product is removed. This version of SPG is called ``SPG+Sequential". Implementation details for both architectures are in Appendix \ref{sec:app-SPG-details}.

\paragraph{Exploration.} We propose to use an exploration strategy inspired by local search methods from meta-heuristic algorithms like the Greedy Randomized Adaptive Search Procedure (GRASP) \cite{Resende2016}. In particular, $k$-exchange neighborhoods are used to perturb the permutations selected by SPG; this exploration heuristic works by randomly swapping $k$ rows of the permutation matrix. In our experiments, we fix $k=2$. This is combined with $\epsilon$-greedy exploration to control the amount of exploration over time. We examine the impact of this exploration strategy on SPG's performance in Section \ref{sec:exploration-study}.

\par 

\section{Experiments}
\label{sec:experiments}

\subsection{Experimental setup}

To evaluate SPG, we considered three combinatorial optimization problems whose solutions can be represented as permutations: sorting with N={20,50}, MWM with N={10,15,20,25}, and the Euclidean TSP with N=20. 

For all experiments, 500K problem instances are generated for training, along with 1K held-out samples for evaluating the target policy. The target policy is evaluated after each pass of the training set by computing the mean reward over the test set. We compute the mean and/or median of the best test set scores over 10 random seeds. For hyperparameter tuning and ablation studies, we used two separate validation set of MWM problem instances with N=10 (MWM-10). Full experiment and hyperparameter details can be found in Appendix \ref{sec:app-hyper}. 

Before presenting results, we will formally introduce the MWM task, which may be less familiar than sorting and the Euclidean TSP. $2N$ points are uniformly sampled in $\mathbb{R}^{[0,1] \times [0,1]}$ to create a bipartite graph $G = (V_1, V_2, E)$, with $\vert V_1 \vert = N$, $\vert V_2 \vert = N$, and $\vert E \vert = N^2$, and each vertex represented by a point $(x_i, y_i)$. The objective is to find an $N \times N$ permutation corresponding to a matching between pairs of vertices from $V_1$ and $V_2$ that maximizes the sum of the Euclidean distances. The action $\mathbf{P}$ selected by the actor is used to permute the vertices of $V_2$ to create a candidate matching. We define an optimality ratio as $\frac{\text{predicted matching weight}}{\text{optimal matching weight}} \in [0, 1]$ to assess performance; optimal matchings can be computed with the Hungarian algorithm. We emphasize that SPG is designed to solve more complex problems than MWM where labels are not easily available (e.g., data association). In these scenarios, SPG must learn to solve the problem from raw inputs and a reward function that may not accurately specify the desired behavior. MWM is useful as a benchmark not only because it is challenging for SPG and the baseline methods, but also because we can exactly evaluate performance through optimality ratios.

\subsection{Ablation and sensitivity studies}
\label{sec:experiments-ablation}
\begin{figure}
\centering
\begin{subfigure}{0.5\textwidth}
\centering
\includegraphics[scale=0.45]{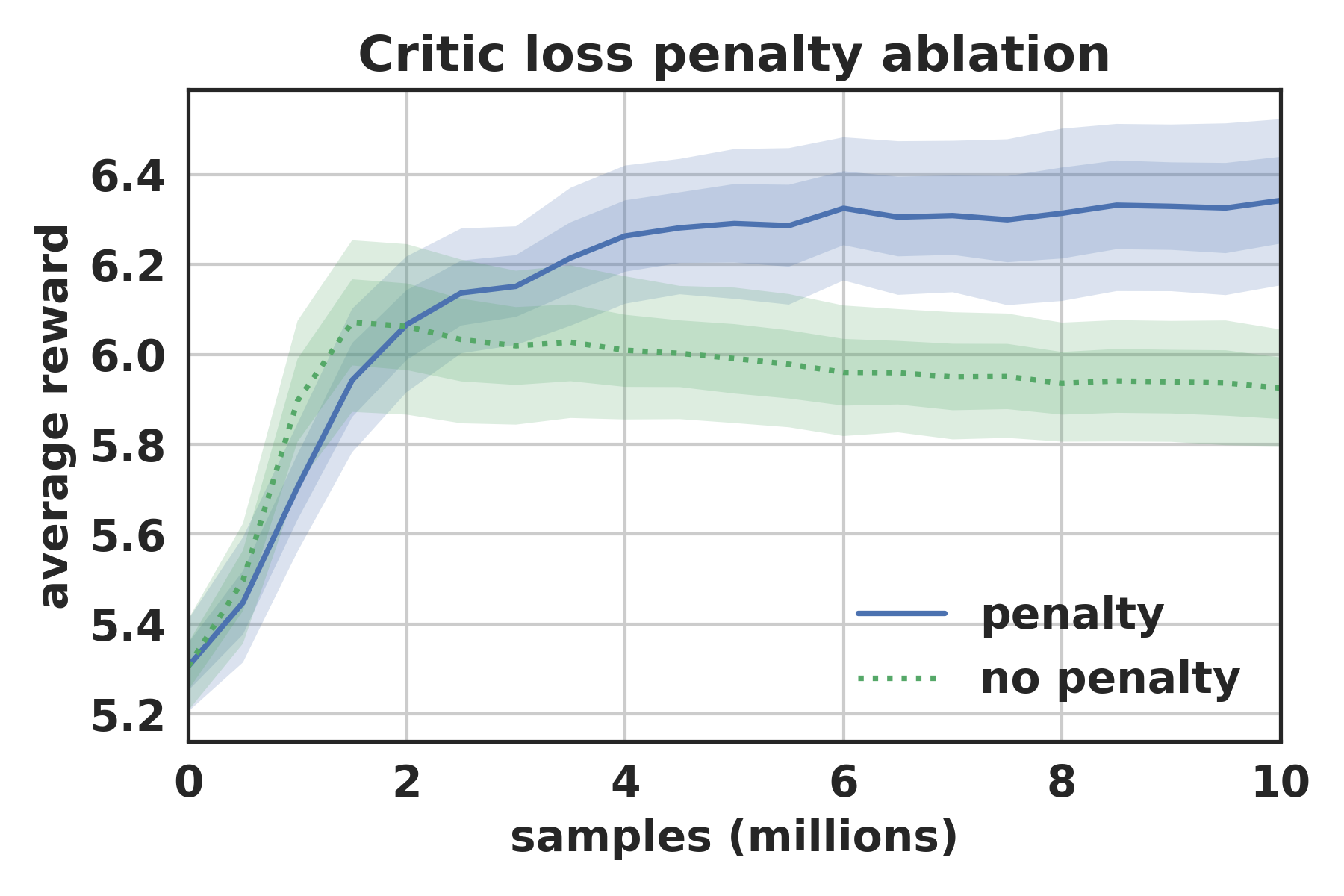}
\caption{\label{fig:critic-penalty-ablation}}
\end{subfigure}%
\begin{subfigure}{0.5\textwidth}
\centering
\includegraphics[scale=0.45]{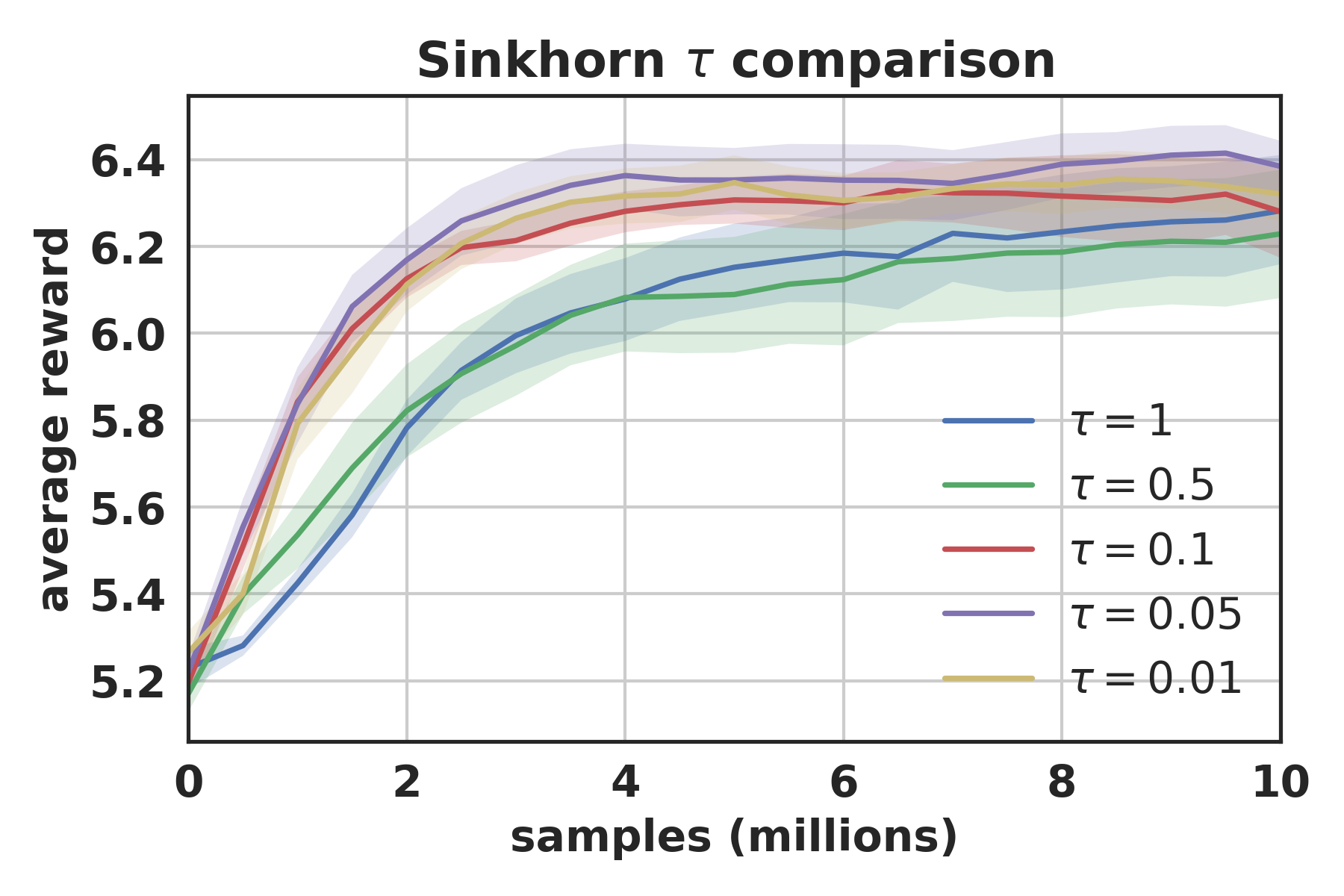}
\caption{\label{fig:sinkhorn-tau-comparison}}
\end{subfigure}
\caption{a) Average reward results on MWM-10 with and without the critic penalty term. With the penalty term, the bias from the continuous relaxation is mostly removed, allowing the actor to keep improving as it sees more data. b) Performance on MWM-10 with various Sinkhorn layer temperatures. We observed that as $\tau$ decreases, the agent can achieve higher average rewards with lower variance. 68\% and 95\% CI bands are shown for (a), and 68\% CI bands are shown for (b).\label{fig:studies}}
\vspace{-1em}
\end{figure}

To improve our understanding of SPG, we conducted the following studies on MWM-10. First, we compared the performance of SPG with and without the added penalty term to the critic loss. Then, we evaluated SPG with and without exploration. Finally, we considered various values for the temperature parameter $\tau$. 
\paragraph{Critic penalty ablation study.}
\label{sec:critic-penalty-ablation}
We performed an ablation study to investigate the effect of the critic loss penalty term. The Q-values for the discrete actions $\mathbf{P}$, which we called ``hard Q-values", accurately predict the immediate rewards. However, without the penalty term, the ``soft Q-values" for continuous actions $\mathbf{M}$ diverge. With the penalty term, both the hard and soft Q-values accurately predict the reward (Figure \ref{fig:penalty-Q-values}, bottom row). The penalty term also allows SPG to train for much longer before learning saturates (Figure \ref{fig:critic-penalty-ablation}).

\paragraph{Exploration study.}
\label{sec:exploration-study}
We evaluated SPG with $\epsilon$-greedy exploration and the $2$-exchange neighborhood heuristic to assess its efficacy. In this experiment, $\epsilon$ is initialized to either one or zero; when initialized to one, we tried linearly decaying $\epsilon$ by $95\%$ and $90\%$ after every epoch until it reached $0.01$. We found that the difference in average reward amongst the considered exploration schemes was insignificant, possibly because the tasks we consider have dense reward functions. However, $\epsilon = 1$ and the decay rate of $0.95$ exhibited the lowest variance for the final average reward after 20 epochs compared to no exploration ($F = 4.402$ with $N_1 = N_2 = 10$ and $p = 0.019$), and hence use this exploration strategy in our experiments. Other starting values for $\epsilon$ between zero and one were considered, but were strictly worse.

\paragraph{Sinkhorn $\tau$ sensitivity analysis.} To ascertain SPG's sensitivity to the Sinkhorn temperature parameter $\tau$, we compared SPG's performance on MWM-10 for values in $\{1, 0.5, 0.1, 0.05, 0.01\}$. Results are displayed in Figure \ref{fig:sinkhorn-tau-comparison}. Smaller values of $\tau$ corresponded to higher average rewards. However, we observed diminishing returns for $\tau < 0.05$, most likely due to increased variance in the policy gradient estimate. We use $\tau = 0.05$ with $L = 10$ in the remainder of the experiments.

\subsection{Main results}
\paragraph{Baselines.} We compare SPG's performance on sorting and the Euclidean TSP against the greedy RL pretraining model from \cite{bello2016neural}. In our experiments, this model is referred to as Pointer-Network Actor-Critic (PN-AC). On the Euclidean TSP, we also compare against \cite{kool2018attention}, an extension of \cite{bello2016neural} that replaces the encoder with a graph attention layer. 

We modify PN-AC to use SPG+Matching's bipartite graph embedding layer for the MWM task; we call this model PN-AC+Matching. We also tried removing PN-AC+Matching's autoregressive decoder to create a simple RL baseline, called AC+Matching. It uses a softmax to sequentially sample match pairs directly from the encoder output. Implementation details for all baseline models are provided in Appendix \ref{sec:app-SPG-details}.   

\paragraph{Sorting.}
SPG is trained to predict the permutation that, when applied to a list of integers, returns them in sorted order.  The reward function is the Kendall-Tau (KT) correlation coefficient, which provides dense rewards. A KT score of $1.0$ means the list was perfectly sorted. We train SPG+Sequential and PN-AC on lists of size $N = \{20, 50\}$. For PN-AC, we use the same hyperparameters as reported in \cite{bello2016neural}, except that we needed to lower the learning rate to 1e-4.
\par
The mean highest attained KT scores on the test set are presented in Table \ref{table:sort-results}. Both SPG and PN-AC are able to learn to solve this task. 

\begin{figure}
\begin{subfigure}[b]{0.33\textwidth}
\includegraphics[scale=0.31]{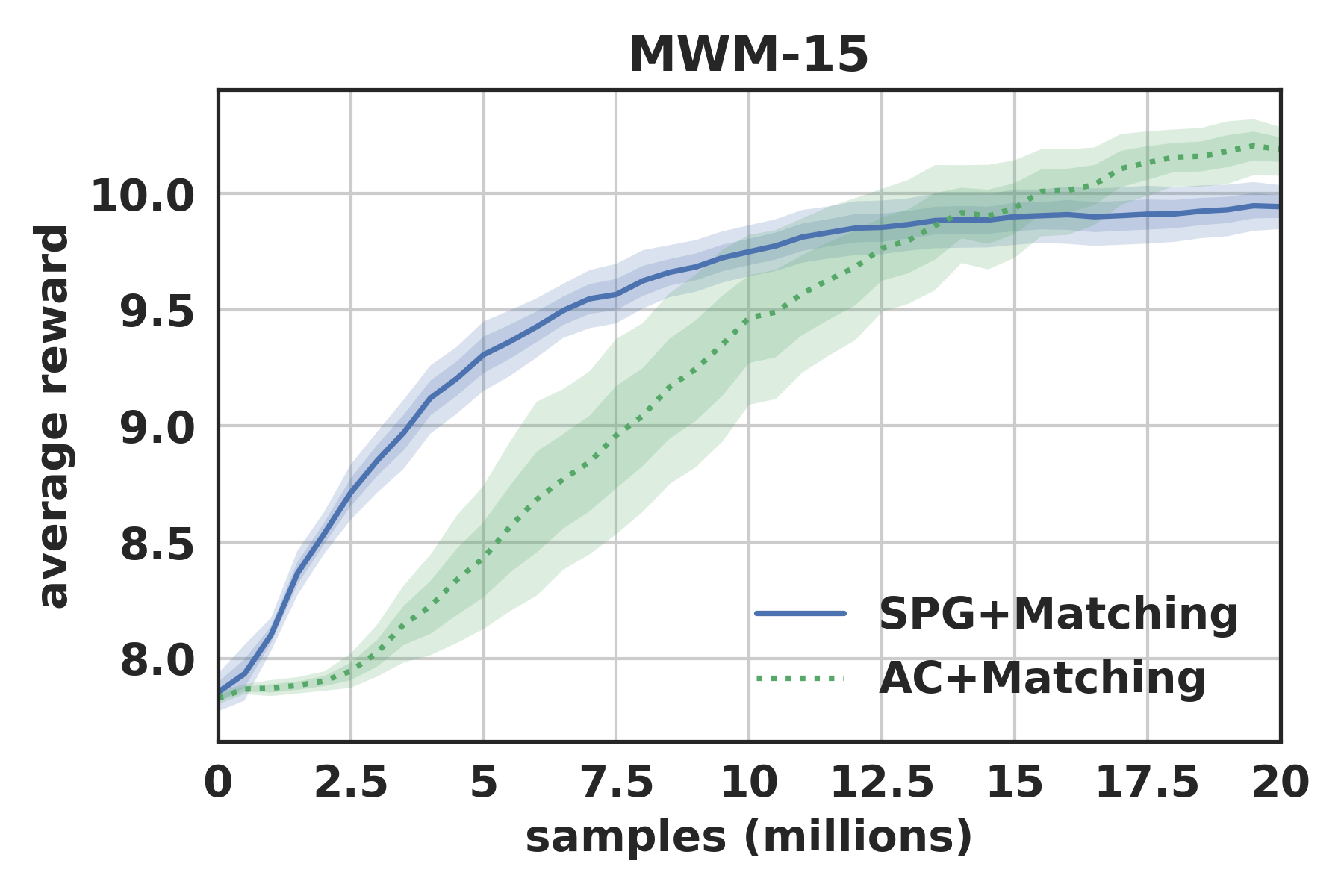}
\caption{}
\end{subfigure}%
\begin{subfigure}[b]{0.33\textwidth}
\includegraphics[scale=0.31]{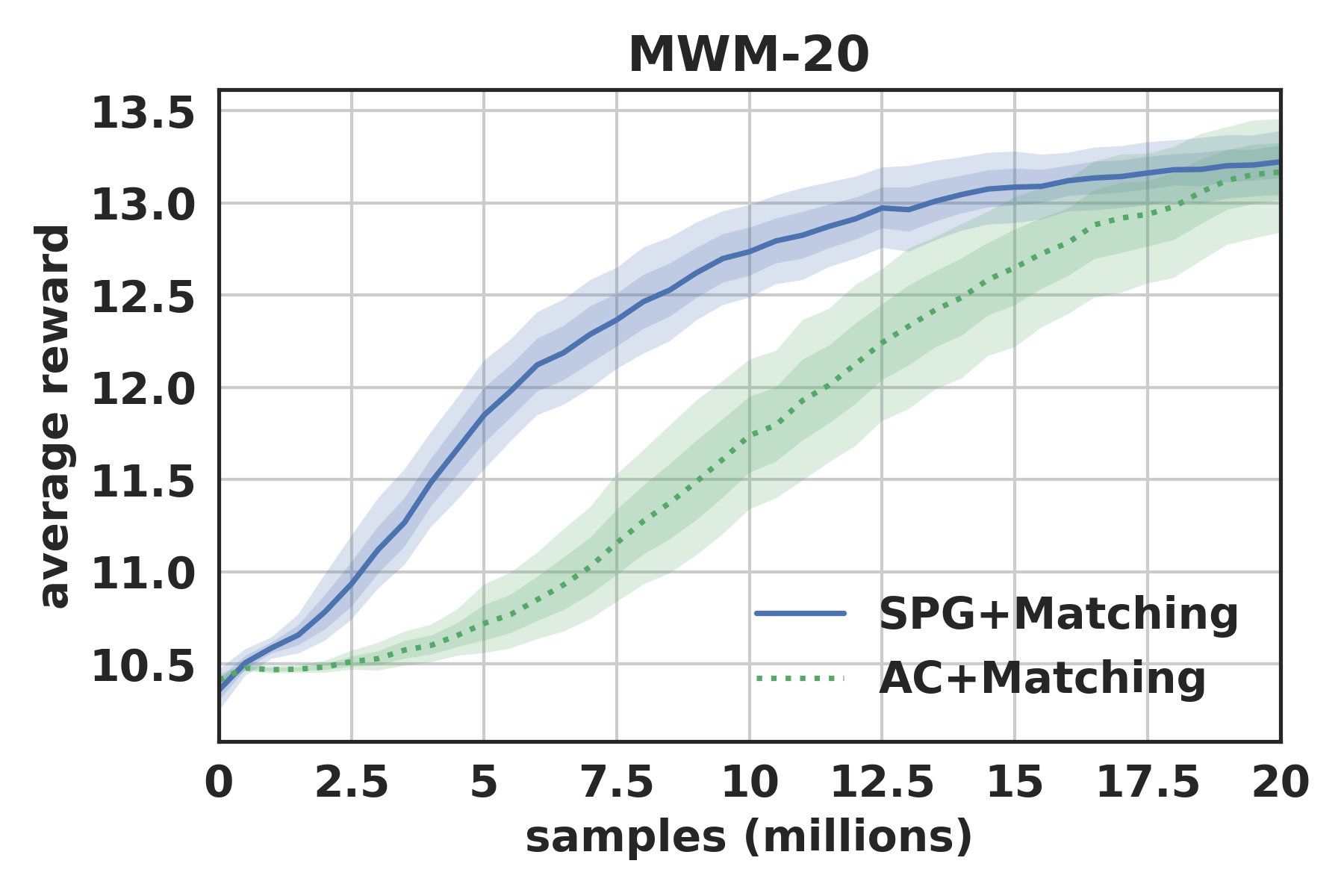}
\caption{}
\end{subfigure}%
\begin{subfigure}[b]{0.33\textwidth}
\includegraphics[scale=0.31]{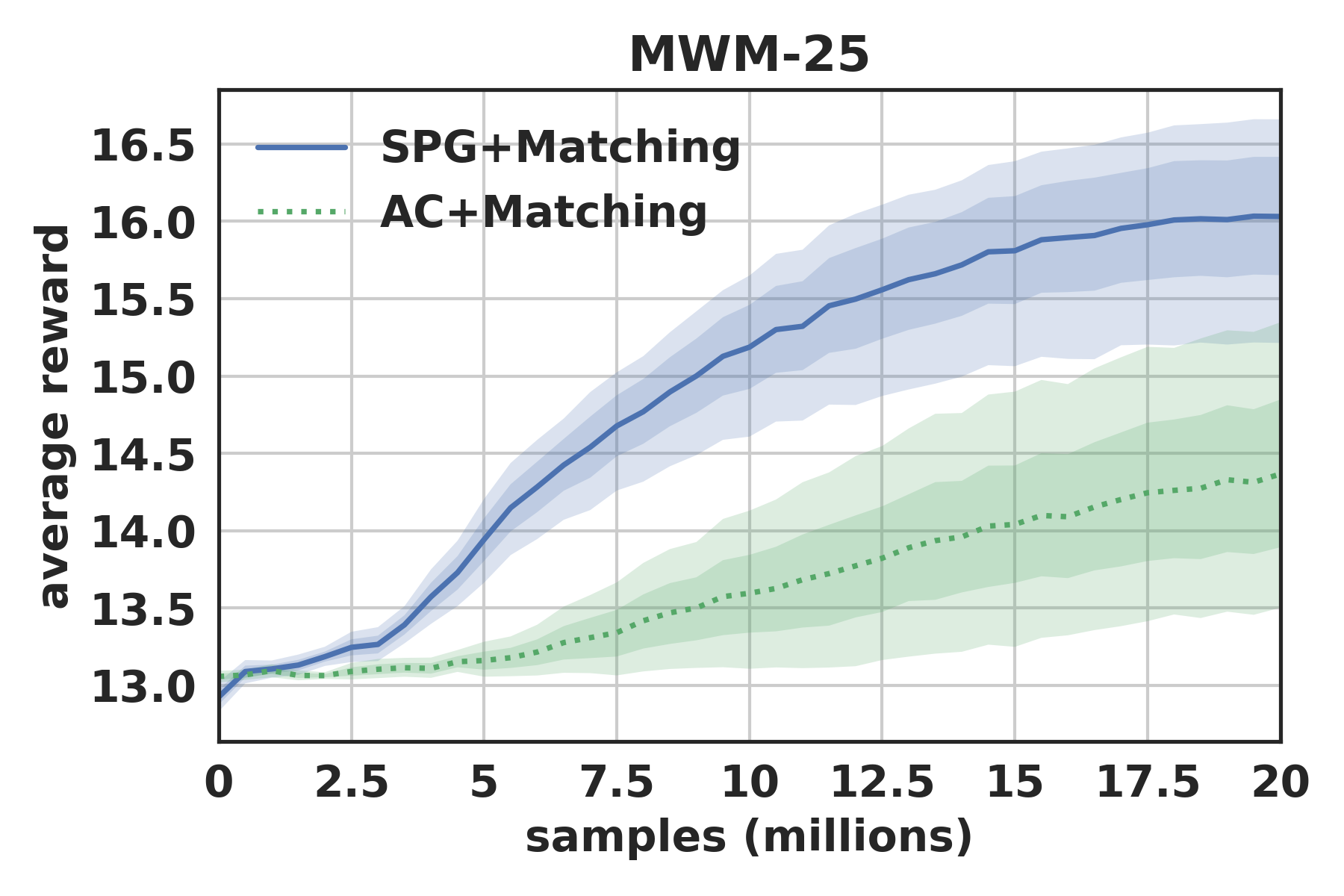}
\caption{}
\end{subfigure}
\caption{SPG+Matching and AC+Matching average reward curves on the MWM test set.  SPG+Matching is more data efficient on all three problem sizes, and achieves a higher average reward on MWM-20 and MWM-25. CI bands for 68\% and 95\% are shown.\label{fig:SPGvsAC}}
\end{figure}

\paragraph{MWM.}
\label{sec:mwm}
For this set of experiments, we trained SPG+Matching, PN-AC+Matching, and AC+Matching on MWM with $N=\{15, 20, 25\}$. As a point of reference, the performance of an untrained SPG+Matching policy  (SPG+Random) is also provided. Results in Table \ref{table:MWM-results} are the medians of the best optimality ratios achieved on the test set over all 10 random seeds. Unlike PN-AC+Matching, AC+Matching is able to do better than random (Figure \ref{fig:SPGvsAC}), from which we conclude that the pointer network decoder is not suitable for this task. SPG+Matching is more data efficient and scales better with larger $N$ than AC+Matching, which implies that the representations learned by SPG+Matching are well-suited to matching problems.

\begin{table}[t]
\begin{center}
\begin{small}
\begin{sc}
\vskip 0.15in
\centering
\tabcolsep=0.11cm
  \begin{minipage}[t]{0.5\textwidth}
  \centering
  \caption{Mean KT on the sorting task.\label{table:sort-results}}
  \vskip 0.15in
  \begin{tabular}{lcccr}
  \toprule
   & N=20 & N=50 \\
  \midrule
  PN-AC& 1.0 $\pm$0 & 0.998$\pm$0.001\\
  SPG+Sequential& 0.998$\pm$0.002 & 0.984$\pm$0.006\\
  \bottomrule
  \end{tabular} 
  \end{minipage}%
  \begin{minipage}[t]{0.5\textwidth}
  \centering
    \caption{Mean tour length on the TSP-20 task.\label{table:tsp-results}}
   \vskip 0.15in
  \begin{tabular}{lc}
  \toprule
  & Avg. Tour Length \\
  \midrule
  Optimal & 3.83 \\
  \cite{kool2018attention} & 3.84 \\
  PN-AC & 3.89 \\
  Christofides & 4.30 \\
  SPG+Sequential & 4.62 \\
  \bottomrule
  \end{tabular}
  \end{minipage}
\end{sc}
\end{small}
\end{center}
\vskip -0.1in
\end{table}

\paragraph{TSP.}
\label{sec:tsp}
To demonstrate SPG's performance on an NP-Hard problem, we trained SPG+Sequential on the Euclidean TSP for $N=20$. To generate each TSP instance, points are uniformly sampled from the unit square. The reward function is the negated sum of the Euclidean distances between each stop along the tour. In Table \ref{table:tsp-results}, we present mean tour lengths on the test. The tour lengths for the baselines are reported from \cite{bello2016neural,kool2018attention}.

\begin{table}
  \caption{Median optimality ratios on the MWM test set.\label{table:MWM-results}}  
  \vskip 0.15in
  \centering
  \begin{tabular}{lcccr}
  \toprule
  & N=15 & N=20 & N=25 \\
  \midrule
  PN-AC+Matching& 0.714 & 0.708 & 0.703\\
  AC+Matching & \textbf{0.935} & \textbf{0.897} & 0.725 \\
  SPG+Random & 0.698 & 0.707 & 0.698\\
  SPG+Matching & 0.904 & \textbf{0.895} & \textbf{0.889}\\
  \bottomrule
  \end{tabular}
\end{table}

\section{Related works}
\label{sec:RW}
The use of learning algorithms for solving problems involving permutations has a rich history. One of the earliest and most influential papers in this line of research, by Hopfield and Tank \cite{hopfield1985neural}, describes how to use Hopfield nets to approximately solve instances of the TSP. Due to the ubiquity of permutations, research on this topic continued, with algorithms based on Sinkhorn balancing and continuous relaxations appearing in \cite{kosowsky1994invisible,gold1996softmax,helmbold2009learning}. The work of \cite{adams2011ranking} laid the groundwork for combining deep learning with learning over permutations \cite{linderman2017reparameterizing,Cruz2017,latentPerms2018}. DeepPermNet, proposed in \cite{Cruz2017}, is a convolutional neural network augmented with a Sinkhorn layer for solving problems in computer vision with supervised learning such as comparing sets of images based on their attributes. The DeepPermNet work is complementary to ours, in that a similar convolutional neural network-based architecture can be trained with SPG to solve permutation-based vision tasks.
 
A supervised-learning approach to the MWM problem is proposed in \cite{milan2017data}. They use a single LSTM to implement a greedy algorithm that sequentially outputs a probability distribution over potential matches one vertex at a time, not unlike the AC+Matching architecture we used as an RL baseline. They show that it can solve a simple data association task for tracking five targets.

The S2V-DQN algorithm from \cite{dai2017learning} uses the \texttt{struct2vec} \cite{dai2016discriminative} algorithm to embed an input graph of a combinatorial problem, which gets passed to a Q-network. For the TSP, we noted that it uses both the edge lengths and well as the vertex coordinates to build its powerful graph representation. PN-AC and SPG both learn their own state representations from just vertex coordinates, and hence both start out producing essentially random tours, unlike S2V-DQN.

\section{Conclusion}
We introduced the SPG algorithm for solving combinatorial optimization problems involving permutations. Our algorithm is able to solve all tasks under consideration reasonably well and demonstrated its ability to learn strong representations for policies over permutations. A current limitation of SPG is that generalizing to different problem sizes requires retraining. This can partially be addressed with an inductive graph-embedding layer \cite{hamilton2017inductive} that removes the embedding layer's dependency on the dimension of the problem size, possibly allowing for positive transfer to larger problem sizes. We note that there are many other ways to potentially improve SPG's performance that are orthogonal to the contributions of this paper, such as incorporating prioritized experience replay \cite{schaul2015prioritized} or parameter noise exploration \cite{plappert2017parameter}.

\bibliography{biblio-nips}

\begin{thebibliography}{37}
\providecommand{\natexlab}[1]{#1}
\providecommand{\url}[1]{\texttt{#1}}
\expandafter\ifx\csname urlstyle\endcsname\relax
  \providecommand{\doi}[1]{doi: #1}\else
  \providecommand{\doi}{doi: \begingroup \urlstyle{rm}\Url}\fi

\bibitem[Adams \& Zemel(2011)Adams and Zemel]{adams2011ranking}
Adams, Ryan~Prescott and Zemel, Richard~S.
\newblock Ranking via sinkhorn propagation.
\newblock \emph{arXiv:1106.1925 [stat.ML]}, 2011.

\bibitem[Bello et~al.(2017)Bello, Pham, Le, Norouzi, and
  Bengio]{bello2016neural}
Bello, Irwan, Pham, Hieu, Le, Quoc~V, Norouzi, Mohammad, and Bengio, Samy.
\newblock Neural combinatorial optimization with reinforcement learning.
\newblock In \emph{Workshop Track of the International Conference on Learning
  Representations}, 2017.

\bibitem[Bengio et~al.(2013)Bengio, L{\'e}onard, and
  Courville]{bengio2013estimating}
Bengio, Yoshua, L{\'e}onard, Nicholas, and Courville, Aaron.
\newblock Estimating or propagating gradients through stochastic neurons for
  conditional computation.
\newblock \emph{arXiv:1308.3432 [cs.LG]}, 2013.

\bibitem[Caetano et~al.(2009)Caetano, McAuley, Cheng, Le, and
  Smola]{caetano2009learning}
Caetano, Tib{\'e}rio~S, McAuley, Julian~J, Cheng, Li, Le, Quoc~V, and Smola,
  Alex~J.
\newblock Learning graph matching.
\newblock \emph{IEEE Transactions on Pattern Analysis and Machine
  Intelligence}, 31\penalty0 (6):\penalty0 1048--1058, 2009.

\bibitem[Cao et~al.(2007)Cao, Qin, Liu, Tsai, and Li]{cao2007learning}
Cao, Zhe, Qin, Tao, Liu, Tie-Yan, Tsai, Ming-Feng, and Li, Hang.
\newblock Learning to rank: from pairwise approach to listwise approach.
\newblock In \emph{Proceedings of the 24th International Conference on Machine
  Learning (ICML'07)}, pp.\  129--136. ACM, 2007.

\bibitem[Cruz et~al.(2017)Cruz, Fernando, Cherian, and Gould]{Cruz2017}
Cruz, Rodrigo~Santa, Fernando, Basura, Cherian, Anoop, and Gould, Stephen.
\newblock Deeppermnet: Visual permutation learning.
\newblock In \emph{Proceedings of the 2017 IEEE Conference on Computer Vision
  and Pattern Recognition (CVPR'17)}, 2017.

\bibitem[Dai et~al.(2016)Dai, Dai, and Song]{dai2016discriminative}
Dai, Hanjun, Dai, Bo, and Song, Le.
\newblock Discriminative embeddings of latent variable models for structured
  data.
\newblock In \emph{Proceedings of the 33rd International Conference on Machine
  Learning (ICML'16)}, pp.\  2702--2711, 2016.

\bibitem[Dai et~al.(2017)Dai, Khalil, Zhang, Dilkina, and
  Song]{dai2017learning}
Dai, Hanjun, Khalil, Elias~B, Zhang, Yuyu, Dilkina, Bistra, and Song, Le.
\newblock Learning combinatorial optimization algorithms over graphs.
\newblock \emph{Advances in Neural Information Processing Systems}, 2017.

\bibitem[Gold \& Rangarajan(1996)Gold and Rangarajan]{gold1996softmax}
Gold, Steven and Rangarajan, Anand.
\newblock Softmax to softassign: Neural network algorithms for combinatorial
  optimization.
\newblock \emph{Journal of Artificial Neural Networks}, 2\penalty0
  (4):\penalty0 381--399, 1996.

\bibitem[Hamilton et~al.(2017)Hamilton, Ying, and
  Leskovec]{hamilton2017inductive}
Hamilton, Will, Ying, Zhitao, and Leskovec, Jure.
\newblock Inductive representation learning on large graphs.
\newblock In \emph{Advances in Neural Information Processing Systems}, pp.\
  1025--1035, 2017.

\bibitem[Helmbold \& Warmuth(2009)Helmbold and Warmuth]{helmbold2009learning}
Helmbold, David~P and Warmuth, Manfred~K.
\newblock Learning permutations with exponential weights.
\newblock \emph{Journal of Machine Learning Research}, 10\penalty0
  (Jul):\penalty0 1705--1736, 2009.

\bibitem[Hopfield \& Tank(1985)Hopfield and Tank]{hopfield1985neural}
Hopfield, John~J and Tank, David~W.
\newblock Neural computation of decisions in optimization problems.
\newblock \emph{Biological cybernetics}, 52\penalty0 (3):\penalty0 141--152,
  1985.

\bibitem[Ioffe \& Szegedy(2015)Ioffe and Szegedy]{ioffe2015batch}
Ioffe, Sergey and Szegedy, Christian.
\newblock Batch normalization: Accelerating deep network training by reducing
  internal covariate shift.
\newblock In \emph{Proceedings of 32nd International Conference on Machine
  Learning (ICML'15)}, pp.\  448--456, 2015.

\bibitem[Jang et~al.(2016)Jang, Gu, and Poole]{jang2016categorical}
Jang, Eric, Gu, Shixiang, and Poole, Ben.
\newblock Categorical reparameterization with gumbel-softmax.
\newblock \emph{arXiv:1611.01144 [stat.ML]}, 2016.

\bibitem[Kingma \& Ba(2014)Kingma and Ba]{kingma2014adam}
Kingma, Diederik and Ba, Jimmy.
\newblock Adam: A method for stochastic optimization.
\newblock \emph{arXiv:1412.6980 [cs.LG]}, 2014.

\bibitem[Kool \& Welling(2018)Kool and Welling]{kool2018attention}
Kool, WWM and Welling, M.
\newblock Attention solves your tsp.
\newblock \emph{arXiv:1803.08475 [stat.ML]}, 2018.

\bibitem[Kosowsky \& Yuille(1994)Kosowsky and Yuille]{kosowsky1994invisible}
Kosowsky, JJ and Yuille, Alan~L.
\newblock The invisible hand algorithm: Solving the assignment problem with
  statistical physics.
\newblock \emph{Neural networks}, 7\penalty0 (3):\penalty0 477--490, 1994.

\bibitem[Kuhn(1955)]{NAV:NAV3800020109}
Kuhn, H.~W.
\newblock The hungarian method for the assignment problem.
\newblock \emph{Naval Research Logistics Quarterly}, 2\penalty0 (1-2):\penalty0
  83--97, 1955.
\newblock ISSN 1931-9193.
\newblock \doi{10.1002/nav.3800020109}.
\newblock URL \url{http://dx.doi.org/10.1002/nav.3800020109}.

\bibitem[Lillicrap et~al.(2015)Lillicrap, Hunt, Pritzel, Heess, Erez, Tassa,
  Silver, and Wierstra]{lillicrap2015continuous}
Lillicrap, Timothy~P, Hunt, Jonathan~J, Pritzel, Alexander, Heess, Nicolas,
  Erez, Tom, Tassa, Yuval, Silver, David, and Wierstra, Daan.
\newblock Continuous control with deep reinforcement learning.
\newblock \emph{arXiv:1509.02971 [cs.LG]}, 2015.

\bibitem[Linderman et~al.(2017)Linderman, Mena, Cooper, Paninski, and
  Cunningham]{linderman2017reparameterizing}
Linderman, Scott~W, Mena, Gonzalo~E, Cooper, Hal, Paninski, Liam, and
  Cunningham, John~P.
\newblock Reparameterizing the birkhoff polytope for variational permutation
  inference.
\newblock \emph{arXiv:1710.09508 [stat.ML]}, 2017.

\bibitem[Maddison et~al.(2017)Maddison, Mnih, and Teh]{maddison2017concrete}
Maddison, Chris~J., Mnih, Andriy, and Teh, Yee~Whye.
\newblock The concrete distribution: A continuous relaxation of discrete random
  variables.
\newblock In \emph{International Conference on Learning Representations}, 2017.

\bibitem[Mena et~al.(2018)Mena, Belanger, Linderman, and
  Snoek]{latentPerms2018}
Mena, Gonzalo, Belanger, David, Linderman, Scott, and Snoek, Jasper.
\newblock Learning latent permutations with gumbel-sinkhorn networks.
\newblock In \emph{International Conference on Learning Representations}, 2018.

\bibitem[Milan et~al.(2017{\natexlab{a}})Milan, Rezatofighi, Dick, Reid, and
  Schindler]{milan2017online}
Milan, Anton, Rezatofighi, Seyed~Hamid, Dick, Anthony~R, Reid, Ian~D, and
  Schindler, Konrad.
\newblock Online multi-target tracking using recurrent neural networks.
\newblock In \emph{Proceedings of the 31st AAAI Conference on Artificial
  Intelligence (AAAI'17)}, pp.\  4225--4232, 2017{\natexlab{a}}.

\bibitem[Milan et~al.(2017{\natexlab{b}})Milan, Rezatofighi, Garg, Dick, and
  Reid]{milan2017data}
Milan, Anton, Rezatofighi, Seyed~Hamid, Garg, Ravi, Dick, Anthony~R, and Reid,
  Ian~D.
\newblock Data-driven approximations to np-hard problems.
\newblock In \emph{Proceedings of the 31st AAAI Conference on Artificial
  Intelligence (AAAI'17)}, pp.\  1453--1459, 2017{\natexlab{b}}.

\bibitem[Mnih et~al.(2015)Mnih, Kavukcuoglu, Silver, Rusu, Veness, Bellemare,
  Graves, Riedmiller, Fidjeland, Ostrovski, et~al.]{mnih2015human}
Mnih, Volodymyr, Kavukcuoglu, Koray, Silver, David, Rusu, Andrei~A, Veness,
  Joel, Bellemare, Marc~G, Graves, Alex, Riedmiller, Martin, Fidjeland,
  Andreas~K, Ostrovski, Georg, et~al.
\newblock Human-level control through deep reinforcement learning.
\newblock \emph{Nature}, 518\penalty0 (7540):\penalty0 529, 2015.

\bibitem[Munkres(1957)]{munkres1957algorithms}
Munkres, James.
\newblock Algorithms for the assignment and transportation problems.
\newblock \emph{Journal of the Society for Industrial and Applied Mathematics},
  5\penalty0 (1):\penalty0 32--38, 1957.

\bibitem[Nowak et~al.(2017)Nowak, Villar, Bandeira, and Bruna]{nowak2017note}
Nowak, Alex, Villar, Soledad, Bandeira, Afonso~S, and Bruna, Joan.
\newblock A note on learning algorithms for quadratic assignment with graph
  neural networks.
\newblock \emph{arXiv:1706.07450 [cs.LG]}, 2017.

\bibitem[Paszke et~al.(2017)Paszke, Gross, Chintala, Chanan, Yang, DeVito, Lin,
  Desmaison, Antiga, and Lerer]{paszke2017automatic}
Paszke, Adam, Gross, Sam, Chintala, Soumith, Chanan, Gregory, Yang, Edward,
  DeVito, Zachary, Lin, Zeming, Desmaison, Alban, Antiga, Luca, and Lerer,
  Adam.
\newblock Automatic differentiation in pytorch.
\newblock 2017.

\bibitem[Plappert et~al.(2017)Plappert, Houthooft, Dhariwal, Sidor, Chen, Chen,
  Asfour, Abbeel, and Andrychowicz]{plappert2017parameter}
Plappert, Matthias, Houthooft, Rein, Dhariwal, Prafulla, Sidor, Szymon, Chen,
  Richard~Y, Chen, Xi, Asfour, Tamim, Abbeel, Pieter, and Andrychowicz, Marcin.
\newblock Parameter space noise for exploration.
\newblock \emph{arXiv:1706.01905 [cs.LG]}, 2017.

\bibitem[Resende \& Ribeiro(2016)Resende and Ribeiro]{Resende2016}
Resende, Mauricio~G.C. and Ribeiro, Celso~C.
\newblock \emph{Optimization by GRASP}.
\newblock Springer New York, 2016.
\newblock \doi{10.1007/978-1-4939-6530-4}.

\bibitem[Schaul et~al.(2015)Schaul, Quan, Antonoglou, and
  Silver]{schaul2015prioritized}
Schaul, Tom, Quan, John, Antonoglou, Ioannis, and Silver, David.
\newblock Prioritized experience replay.
\newblock \emph{arXiv:1511.05952 [cs.LG]}, 2015.

\bibitem[Silver et~al.(2014)Silver, Lever, Heess, Degris, Wierstra, and
  Riedmiller]{silver2014deterministic}
Silver, David, Lever, Guy, Heess, Nicolas, Degris, Thomas, Wierstra, Daan, and
  Riedmiller, Martin.
\newblock Deterministic policy gradient algorithms.
\newblock In \emph{Proceedings of the 31st International Conference on Machine
  Learning (ICML'14)}, pp.\  387--395, 2014.

\bibitem[Sinkhorn(1964)]{sinkhorn1964relationship}
Sinkhorn, Richard.
\newblock A relationship between arbitrary positive matrices and doubly
  stochastic matrices.
\newblock \emph{The Annals of Mathematical Statistics}, 35\penalty0
  (2):\penalty0 876--879, 1964.

\bibitem[Vinyals et~al.(2015{\natexlab{a}})Vinyals, Bengio, and
  Kudlur]{vinyals2015order}
Vinyals, Oriol, Bengio, Samy, and Kudlur, Manjunath.
\newblock Order matters: Sequence to sequence for sets.
\newblock \emph{arXiv:1511.06391 [stat.ML]}, 2015{\natexlab{a}}.

\bibitem[Vinyals et~al.(2015{\natexlab{b}})Vinyals, Fortunato, and
  Jaitly]{vinyals2015pointer}
Vinyals, Oriol, Fortunato, Meire, and Jaitly, Navdeep.
\newblock Pointer networks.
\newblock In \emph{Advances in Neural Information Processing Systems}, pp.\
  2692--2700, 2015{\natexlab{b}}.

\bibitem[Williams(1992)]{williams1992simple}
Williams, Ronald~J.
\newblock Simple statistical gradient-following algorithms for connectionist
  reinforcement learning.
\newblock In \emph{Reinforcement Learning}, pp.\  5--32. Springer, 1992.

\bibitem[Wojke et~al.(2017)Wojke, Bewley, and Paulus]{Wojke2017simple}
Wojke, Nicolai, Bewley, Alex, and Paulus, Dietrich.
\newblock Simple online and realtime tracking with a deep association metric.
\newblock In \emph{2017 IEEE International Conference on Image Processing
  (ICIP)}, pp.\  3645--3649, 2017.

\end{thebibliography}
\bibliographystyle{icml2018}

\appendix

\section{Actor-critic architectures}

\label{sec:app-SPG-details}

\subsection{SPG+Matching}
We begin by explaining the SPG+Matching actor network architecture for $\mathcal{S} = \mathbb{R}^{N \times K} \times \mathbb{R}^{N \times K}$. Let $N$ be the number of objects in each of the two disjoint sets of vertices in a bipartite graph and suppose $\mathbf{X}_1, \mathbf{X}_2 \in \mathbb{R}^{N \times K}$ is a representation of the bipartite graph with feature dimension $K$. For planar graphs, as in MWM, $K = 2$. The parameters for the nonlinear embedding of $\mathbf{X}_1$ and $\mathbf{X}_2$, $\{\mathbf{W}_e, \mathbf{b}_e\}$ are shared,

\begin{equation}
\label{eq:subgraph-embed}
\begin{aligned}
\mathbf{E}_1 &= \sigma \big(\mathbf{X}_1 \mathbf{W}_e + \mathbf{b}_e\big) \\
\mathbf{E}_2 &= \sigma \big(\mathbf{X}_2 \mathbf{W}_e + \mathbf{b}_e\big) 
\end{aligned}
\end{equation}

with $\mathbf{W}_e \in \mathbb{R}^{K \times 128}$ and $\mathbf{b}_e \in \mathbb{R}^{128}$. All nonlinearities $\sigma$ are LeakyReLUs with negative slope $0.01$. Next, the matrix outer product of the embeddings is computed

\begin{equation}
\label{eq:fused-embed}
\mathbf{E} = \mathbf{E}_2 \mathbf{E}_1^\intercal.
\end{equation}

$\mathbf{E}$ is split in the zeroth dimension to form an $N$-dimensional sequence of length $N$. Next, this sequence gets passed to a GRU layer:

\begin{equation}
\mathbf{h}_N = \text{GRU}(\mathbf{E}, \mathbf{h}_0).
\end{equation}

The GRU has parameters $\mathbf{W}_i \in \mathbb{R}^{3N \times 128}, \mathbf{b}_i \in \mathbb{R}^{3 * 128}$, $\mathbf{W}_h \in \mathbb{R}^{3N \times 128}, \mathbf{b}_h \in \mathbb{R}^{3 * 128}$, and the output of the GRU is $\mathbf{h}_N \in \mathbb{R}^{N \times 128}$.

We map $\mathbf{h}_N$ to an $N \times N$ matrix with a linear layer, 

\begin{equation}
\label{eq:app-actor-out}
\mathbf{Y}_a = \mathbf{h}_N \mathbf{W}_a + \mathbf{b}_a
\end{equation}

for $\mathbf{W}_a \in \mathbb{R}^{128 \times N}$ and $\mathbf{b}_a \in \mathbb{R}^{N}$. Next, $\mathbf{Y}_a$ is processed by the Sinkhorn layer (Figure \ref{fig:code}); in our experiments, we use $\tau = 0.05$ and $L = 10$. The Hungarian algorithm is applied to round the output of the Sinkhorn layer to a permutation matrix. We parallelize the Hungarian algorithm by splitting the mini-batch across multiple processors. 

The critic network for SPG+Matching has a similar architecture as the actor network up to Equation \ref{eq:app-actor-out}. Instead of the linear layer, we map the hidden state of the GRU back to the embedding dimension with a nonlinear layer and BatchNorm \cite{ioffe2015batch},

\begin{equation}
\mathbf{Y}_c = \sigma \big(\text{BN}(\mathbf{h}_N \mathbf{W}_c + \mathbf{b}_c) \big).
\end{equation}

for $\mathbf{W}_c \in \mathbb{R}^{128 \times 128}$ and $\mathbf{b}_c \in \mathbb{R}^{128}$. The critic also takes as input the action $\mathbf{P}$, which we combine with the learned representation of the state $\mathbf{Y}_c$ as follows:

\begin{equation}
\begin{aligned}
\mathbf{E}_d &= \sigma \big(\text{BN}(\mathbf{P} \mathbf{W}_d + \mathbf{b}_d )\big) \\
\mathbf{Y}_f &= \sigma \big(\text{BN} \big((\mathbf{Y}_c + \mathbf{E_d} ) \mathbf{W}_f + \mathbf{b}_f \big) \big).
\end{aligned}
\end{equation}

with $\mathbf{W}_d \in \mathbb{R}^{N \times 128}$ and $\mathbf{b}_d \in \mathbb{R}^{128}$, and $\mathbf{W}_f \in \mathbb{R}^{128 \times N}$ and $\mathbf{b}_f \in \mathbb{R}^{N}$. We found that BatchNorm helps the critic network combine the state and action effectively. The fused state and action embedding $\mathbf{Y}_f \in \mathbb{R}^{N \times N}$ is mapped to a scalar by 

\begin{equation}
Q = (\mathbf{Y}_f \mathbf{W}_{g_1})^{\intercal} \mathbf{W}_{g_2}
\end{equation}

where $\mathbf{W}_{g_1}, \mathbf{W}_{g_2} \in \mathbb{R}^{N \times 1}$.

\subsection{SPG+Sequential}

The SPG+Sequential architecture is used for problems where $\mathcal{S} = \mathbb{R}^{N \times K}$ is a set of $N$ objects, and the agent selects an action that permutes its own input. For the most part, the SPG+Sequential actor network mirrors the actor architecture from SPG+Matching. Rather than reproduce most of Equations 8-12 here, we will just point out the few differences. First, the SPG+Sequential actor only needs to embed one set of objects $\mathbf{X}$, so there is only one embedding computed instead of two (Equation 8). Second, this embedding is passed directly to the GRU; there is no matrix outer product as in Equation 9. The GRU in SPG+Sequential is bidirectional, which we found helped improve the overall performance on TSP-20.
\par
SPG+Sequential's critic network differs from SPG+Matching's critic network in that the action is fused with the state \textit{before} the recurrent layer. This performed significantly better for the tasks we considered; we believe that this is due to the prominent sequential nature of both the states and actions for the sorting and TSP tasks. The critic can be defined for a state and action pair $(\mathbf{X}, \mathbf{P})$ as

\begin{align}
& \mathbf{E}_1 = \sigma \big( \text{BN} ( \mathbf{X} \mathbf{W}_j + \mathbf{b}_j) \big) \\
& \mathbf{E}_2 = \sigma \big( \text{BN} (\mathbf{P} \mathbf{W}_k + \mathbf{b}_k) \big) \\
& \mathbf{E} = \sigma \big( \text{BN} \big(\mathbf(\mathbf{E}_1 + \mathbf{E}_2) \mathbf{W}_l + \mathbf{b}_l \big) \big) \\
& \mathbf{h}_N = \text{GRU}(\mathbf{E}, \mathbf{h}_0) \\
& \mathbf{Y}_m = \sigma ( \mathbf{h}_N \mathbf{W}_m + \mathbf{b}_m ) \\
& Q = (\mathbf{Y}_m \mathbf{W}_{n_1})^{\intercal} \mathbf{W}_{n_2}.
\end{align}

The critic parameters are $\mathbf{W}_j \in \mathbb{R}^{K \times 128}, \mathbf{b}_j \in \mathbb{R}^{128}, \mathbf{W}_k \in \mathbb{R}^{N \times 128}, \mathbf{b}_k \in \mathbb{R}^{128}, \mathbf{W}_l \in \mathbb{R}^{128 \times 128}, \mathbf{b}_l \in \mathbb{R}^{128}, \mathbf{W}_m \in \mathbb{R}^{256 \times 128}, \mathbf{b}_m \in \mathbb{R}^{128}, \mathbf{W}_{n_1} \in \mathbb{R}^{128 \times 1}, \mathbf{W}_{n_2} \in \mathbb{R}^{128 \times 1}$, plus the GRU parameters with dimension 256 (2x 128 for bidirectional GRU).

\paragraph{Sample code for implementing a Sinkhorn layer.}
In Figure \ref{fig:code}, we show example code for a stable implementation of the Sinkhorn layer. In our experimentation, a naive implementation occasionally caused numerical errors even for $\tau$ values as large as 0.1. 

\subsection{PN-AC}
For the sorting task, we used an implementation of PN-AC that is based on the greedy RL pretraining model from \cite{bello2016neural}. We briefly describe the architecture here, but refer the reader to \cite{bello2016neural} for more details. The actor network in PN-AC uses an LSTM encoder and decoder with an attention mechanism to sequentially ``point" at elements from the input sequence to greedily construct solutions. This algorithm is able to produce solutions that satisfy the permutation constraints by hard-coding a rule that prevents the attention mechanism from selecting the same element from the input twice. Instead of a critic network, we found that an exponential moving average made a suitable baseline for PN-AC's REINFORCE policy gradient estimator. Crucially, PN-AC is designed to perform well on combinatorial problems where the output is a permutation or subset of the input, and both the input and output have an underlying sequential structure. 

\subsection{PN-AC+Matching}

We modified the PN-AC architecture to train it on the MWM task. In PN-AC, the length $N$ input sequence $\mathbf{X} \in \mathbb{R}^{N \times K}$ representing the state gets embedded by a linear transformation and then processed by an encoder network. PN-AC+Matching instead adopts the shared dual-embedding of SPG+Matching to transform two inputs $\mathbf{X}_1, \mathbf{X}_2 \in \mathbb{R}^{N \times K}$ into the embedding $\mathbf{E} \in \mathbb{R}^{N \times N}$ from Equation \ref{eq:fused-embed}. As in SPG+Matching, $\mathbf{E}$ is split in the zeroth dimension into a length $N$ sequence of $N$-dimensional vectors, which gets processed by an identical encoder network to the one used by PN-AC. The output of the encoder, a length $N$ sequence of 128-dimensional vectors, is passed to a decoder network as the attention context. The decoder network is mostly unchanged from the one used by PN-AC. As with PN-AC, the initial decoder input is a trainable parameter. We concatenate the embeddings $\mathbf{E}_1$ and $\mathbf{E}_2$ from Equation \ref{eq:subgraph-embed} to use as inputs to the decoder, and hence the decoder input dimension is 256. At each time step $t = 1, ..., N$, the pointer mechanism selects the i\textsuperscript{th} node, $i = 1, ..., N$, from $\mathbf{X}_2$. The $i$\textsuperscript{th} element of $\mathbf{E}_2$ is concatenated with the $t$\textsuperscript{th} element of  $\mathbf{E}_1$ to use as the input to the decoder at the next time step. The order in which the elements of $\mathbf{E}_2$ are selected dictates the permutation applied to $\mathbf{X}_2$ after the decoding is complete, which is used to form the matching solution.  
\par
In summary, the main difference between PN-AC and PN-AC+Matching is the embedding process for the input, which is identical to the one used by SPG+Matching. 

\subsection{AC+Matching}

To create a simpler (albeit more effective) baseline than PN-AC+Matching, we experimented with removing the autoregressive decoder. This architecture also uses the same embedding process as PN-AC+Matching and SPG+Matching. Instead of using a decoder, it simply takes the fused embedding $\mathbf{E}$, splits it along the zeroth dimension, and uses each $N$-dimensional vector as input to an LSTM. At each time step, the output of the LSTM $\mathbf{h}_t \in \mathbb{R}^{128}$ is transformed by

\begin{equation}
\mathbf{l}_t = C \tanh(\mathbf{h}_t \mathbf{W}_p + \mathbf{b}_p),
\end{equation}

where $C = 10$, $\mathbf{W}_p \in \mathbb{R}^{128 \times N}$, and $\mathbf{b}_p \in \mathbb{R}^{N}$. The logits $\mathbf{l}_t \in \mathbb{R}^N$ are masked, as is done in PN-AC and PN-AC+Matching, to keep inputs from being selected more than once by the pointer mechanism. A normalized softmax operation is applied to $l$ to create a multinomial distribution over the available inputs, from which one input element is stochastically selected. Similarly to PN-AC and PN-AC+Matching, AC+Matching is trained with REINFORCE and uses an exponential moving-average baseline.

\lstset{
  backgroundcolor=\color{white},
  basicstyle=\fontsize{7.5pt}{8.5pt}\fontfamily{lmtt}\selectfont,
  columns=fullflexible,
  breaklines=true,
  captionpos=b,
  commentstyle=\fontsize{8pt}{9pt}\color{codegray},
  keywordstyle=\fontsize{8pt}{9pt}\color{codegreen},
  stringstyle=\fontsize{8pt}{9pt}\color{codeblue},
  frame=tb,
  otherkeywords = {self},
}
\begin{figure}[t]
\tiny
\begin{lstlisting}[language=python]
def SinkhornLayer(x, tau, L, eps=1e-6):
    # x: input features with shape [N,N]
    # tau: temperature parameter
    # L: number of Sinkhorn iters
    
    x = x / tau
    for _ in range(L):
    	# row normalization
    	x = x - LogSumExp(x, dim=1, keepdims=True)
        # column normalization
        x = x - LogSumExp(x, dim=0, keepdims=True)
    
    # add a small offset 'eps' to avoid numerical 
    # errors due to exp()
    return exp(x) + eps
\end{lstlisting}
\caption{Generic Python code for a stable implementation of the Sinkhorn layer. The \texttt{dim} argument of \texttt{LogSumExp} indicates which axis to sum over.}
\label{fig:code}
\vspace{-1em}
\end{figure}

\section{Algorithm}
\label{sec:app-alg}
\begin{algorithm}[tb]
   \caption{Sinkhorn Policy Gradient}
   \label{alg:SPG}
\begin{algorithmic}
   \STATE Initialize actor $\pi_\theta(s; \tau)$ and critic $Q_{\theta'}(s,a)$\\
   \STATE Initialize replay buffer $R$
   \FOR{$i=1$ {\bfseries to} $max\_train\_steps$}
   \STATE Sample state $s \sim \rho$
   \STATE $\mathbf{M} = \pi_\theta(s; \tau)$
   \STATE $\mathbf{P} = \text{H}(\mathbf{M})$
   \STATE Sample $u \sim \text{Uniform}[0,1)$
   \IF{$u < \epsilon$}
     \STATE Make $k=2$ random row exchanges for $\mathbf{P}$ and $\mathbf{M}$ 
   \ENDIF
   \STATE Apply $\mathbf{P}$ to $s$ and observe $r(s,\mathbf{P})$
   \STATE Store experience ($s$, $\mathbf{M}$, $\mathbf{P}$, $r$) in $R$
   \STATE Sample mini-batch ($s_n, \mathbf{M}_n,\mathbf{P}_n, r_n)$ $\sim R$
   \STATE Update critic by minimizing: 
   \begin{align*}
   	\text{MSE}\big(r_n, Q_{\theta'}(s_n, \mathbf{P}_n)\big)
    + \text{MSE}\big( \texttt{stop\_grad}(Q_{\theta'}(s_n, \mathbf{P}_n)), Q_{\theta'}(s_n, \mathbf{M}_n)\big)
   \end{align*}
   \STATE $\mathbf{M_n'} = \pi_\theta(s_n; \tau)$
   \STATE Update the actor policy by ascending the sampled policy gradient:
   \begin{align*}
   \nabla_\theta & \pi_\theta \approx \frac{1}{N} \sum_n \nabla_\theta \pi_\theta (s_n; \tau) \nabla_{a} Q_{\theta'}(s_n, a) \big \vert_{a=\mathbf{M_n'}}
   \end{align*}
   \ENDFOR
\end{algorithmic}
\end{algorithm}

The training algorithm we use for SPG is given in Algorithm \ref{alg:SPG}. A replay buffer $R$ is maintained for experience replay; we uniformly sample from $R$ to construct mini-batches for computing gradient estimations. Each epoch lasts for a pre-determined number of training steps, after which the target policy is evaluated on a held-out test set.

\section{Geometric interpretation of the critic loss penalty term}
\label{sec:app-critic-penalty}
\begin{figure}
\centering
\begin{subfigure}[t]{0.45\textwidth}
\includegraphics[scale=0.28]{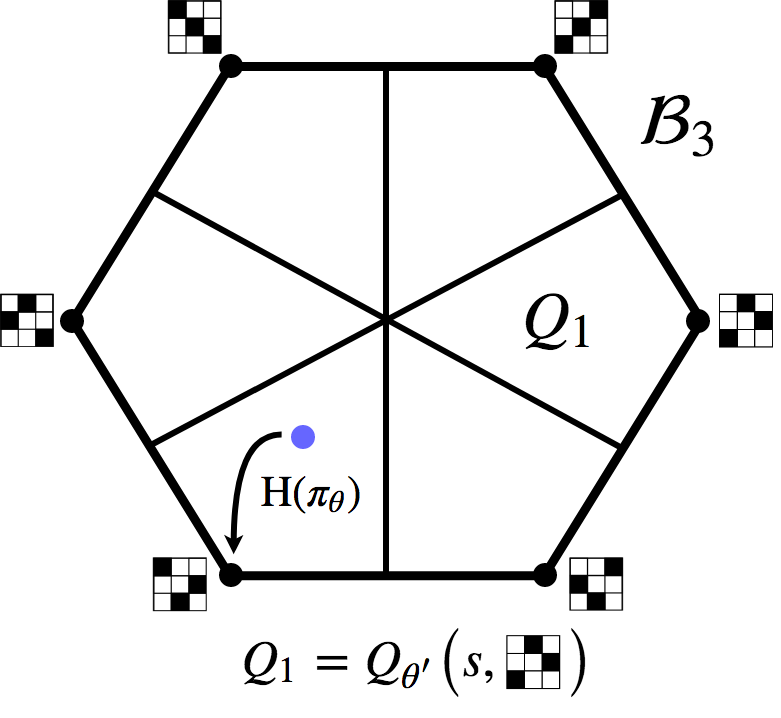}
\caption{\label{fig:B3}}
\end{subfigure}%
\begin{subfigure}[t]{0.55\textwidth}
\includegraphics[scale=0.155]{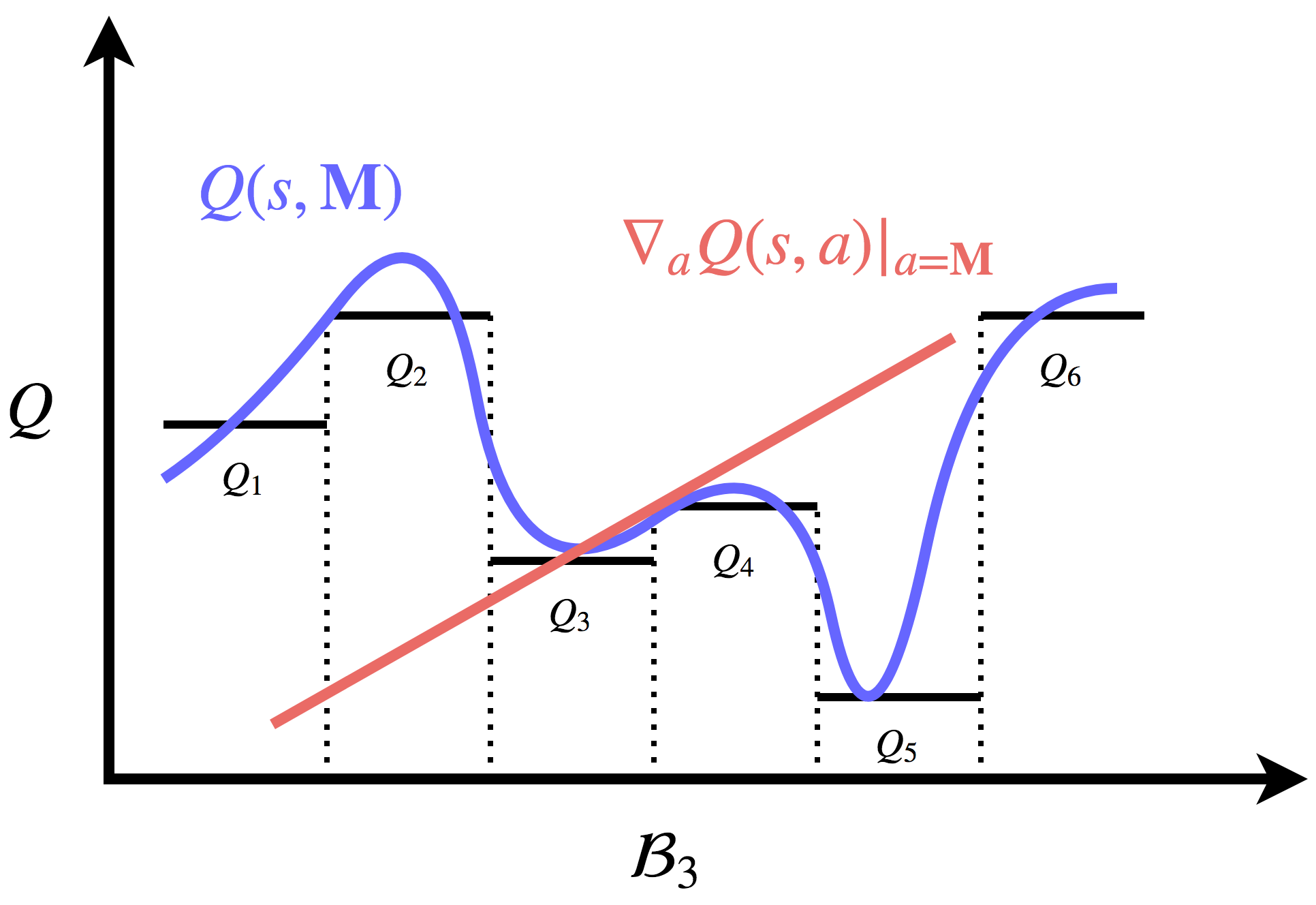}
\caption{\label{fig:BNvQ}}
\end{subfigure}
\caption{a) Visual representation of the Birkhoff polytope with $N = 3$ ($\mathcal{B}_3$) as a hexagon, for some fixed $s \in \mathcal{S}$. The points in $\mathcal{B}_3$ are $3 \times 3$ doubly-stochastic matrices; the vertices of the $\mathcal{B}_3$/hexagon are $3 \times 3$ permutation matrices. All elements of $\mathcal{B}_3$ within a region are sent to the vertex contained in that region by the Hungarian method, and hence have the same Q-value, e.g., $Q_1$. b) In this figure, the x-axis is divided into the 6 regions of $\mathcal{B}_3$, as if the hexagon from (a) was cut to the center along one of the region boundaries and laid out along the real line. This is not a technically accurate depiction of the Q-values over $\mathcal{B}_3$, but it is useful to acquire intuition. The Q-values for discrete actions, i.e. permutations, are piece-wise constant ($Q_{1-6}$) and the Q-values for the continuous actions form a smooth, differentiable surface. The penalty term tries to minimize the squared error between the two so that the action-gradient of the smooth surface provides an accurate indication of the direction of maximum reward improvement at region boundaries.\label{fig:critic-loss-proof}}
\end{figure}

In this section, we provide a geometric interpretation of the penalty term used to remove bias due to the continuous relaxation. First, we define the \textit{Birkhoff polytope} $\mathcal{B}_N$ as the set of all $N \times N$ doubly-stochastic  matrices. All points of $\mathcal{B}_N$ lie on the convex hull of $\mathcal{P}_N$, i.e., the vertices of the convex hull are $N \times N$ permutations. Therefore, an SPG policy $\pi_{\theta}$ outputs an element $\mathbf{M} \in \mathcal{B}_N$, which gets mapped to the nearest vertex of $\mathcal{B}_N$ by the non-differentiable Hungarian method, i.e., $\mathbf{P} = \text{H}(\pi_{\theta})= \text{H}(\mathbf{M})$. We provide a visual representation of $\mathcal{B}_3$ in Figure \ref{fig:B3} as an example. The Q-values for the discrete actions $Q_{\theta'}(s, \mathbf{P})$ are piece-wise constant over $\mathcal{B}_N$, which is why $\nabla_{a} Q_{\theta'}(s,a) \big \vert_{a = \mathbf{P}}$ is zero almost everywhere (and undefined at the discontinuities between the regions dividing $\mathcal{B}_N$). Next, we will describe how we can obtain a useful critic action-gradient that accurately approximates the direction of maximum reward improvement.

We can measure how close the Q-values for the relaxed permutations $\mathcal{M}$ match those for $\mathcal{P}$ with the following: 

\[\epsilon = \int_{\mathcal{S}} \big( Q_{\theta'}\big(s, \mathbf{P}\big) - Q_{\theta'}(s,\mathbf{M}) \big)^2 \text{d}s.\]

Figure \ref{fig:BNvQ} shows the Q-values corresponding to $\mathbf{P}$ and $\mathbf{M}$ for fixed $s \in \mathcal{S}$ in $\mathcal{B}_3$. By treating this as a regression problem with loss function $\epsilon$ and targets $Q_{\theta'}\big(s, \mathbf{P}\big)$, we fit a smooth approximation to the piece-wise constant action-value function induced by $\mathbf{P}$. The critic action-gradient $\nabla_{a} Q_{\theta'}(s,a) \big \vert_{a = \mathbf{M}}$, depicted in Figure \ref{fig:BNvQ}, is well-defined everywhere in $\mathcal{B}_N$ since $Q_{\theta'}(s, \mathbf{M})$ is continuous and differentiable by construction. Furthermore, the critic action-gradient clearly approximates the direction of reward improvement at the region boundaries in $\mathcal{B}_N$, and the quality of the approximation improves as $\epsilon \rightarrow 0$.  

In practice, we can minimize $\epsilon$ by minimizing the mean squared error between $Q_{\theta'}(s, \mathbf{P})$ and $Q_{\theta'}(s, \mathbf{M})$, which we estimate with a mini-batch of samples $s$ drawn from $\rho$ as $\text{MSE} \big( \texttt{stop\_grad}\big( Q_{\theta'}(s, \mathbf{P}) \big) , Q_{\theta'}(s, \mathbf{M}) \big)$. Updating $\theta'$ with the gradient of $\epsilon$ and a small enough step size guarantees that $\epsilon$ will decrease and eventually settle in some local minima. Even though the targets use the same parameters as $Q_{\theta'}(s, \mathbf{M})$, we hold them fixed with \texttt{stop\_grad} and we observe that they become stable as the critic converges.

\section{Experiment details}
\label{sec:app-hyper}
Details about the deep network architectures for SPG and the baselines are provided in Appendix \ref{sec:app-SPG-details}. We use Adam \cite{kingma2014adam} to optimize all models. We arrived at the hyperparameters listed in Table \ref{table:hyperparams} using the SigOpt Bayesian optimization service (except for Adam's parameters, which are the defaults) on a validation set of 500K instances of MWM-10. These hyperparameters are shared across all tasks in our experiments for both SPG+Matching and SPG+Sequential.
\par
For the number of Sinkhorn iterations $L$, we considered $L \in \{5, 10, 15\}$ and found that 10 offered the best trade-off between having too few iterations (and hence the matrix $\mathbf{M}$ doesn't satisfy the permutation sum-to-one constraints for both the rows and columns), or too many iterations which causes the gradients to vanish.

All experiments are run with pure PyTorch \cite{paszke2017automatic} and a single NVIDIA GPU (we used a GTX 1080 and a server with Titan Xp's and a Titan V). We note that we observed a significant speed up when splitting the Hungarian algorithm computation over across multiple cores. In our implementation, we split batches of problem instances of size 128 across four cores. Code for reproducing the results from this paper will be available online. 

\begin{table}[t]
\caption{Hyperparameters used in our main experiments.}
\label{table:hyperparams}
\vskip 0.15in
\begin{center}
\begin{small}
\begin{sc}
\tabcolsep=0.11cm
\begin{tabular}{lcccr}
\toprule
 Name & Value \\
\midrule
Optimization \\
\midrule
actor LR & 1e-5 \\
actor LR decay rate & 5\%/5K steps \\
critic LR & 2e-4 \\
critic LR decay rate & 5\%/5K steps \\
Adam $\beta_1$ & 0.9 \\
Adam $\beta_2$ & 0.999 \\
Adam $\epsilon$ & 1e-8 \\
\midrule
Exploration \\
\midrule
$k$ & 2 \\
$\epsilon$-start & 1.0 \\
$\epsilon$-end & 0.01 \\
$\epsilon$-decay & 5\%/epoch \\
\midrule
Sinkhorn layer \\
\midrule
$\tau$ & 0.05 \\
$L$ & 10 \\
\midrule 
Training \\
\midrule
mini-batch size & 128 \\
replay buffer size & 1e6 \\
L2 gradient norm clipping & 1 \\
\bottomrule
\end{tabular}
\end{sc}
\end{small}
\end{center}
\vskip -0.1in
\end{table}

\section{Analysis}
\label{sec:app-analysis}
\begin{figure}
\centering
\begin{subfigure}[t]{.4\textwidth}
\includegraphics[scale=0.35]{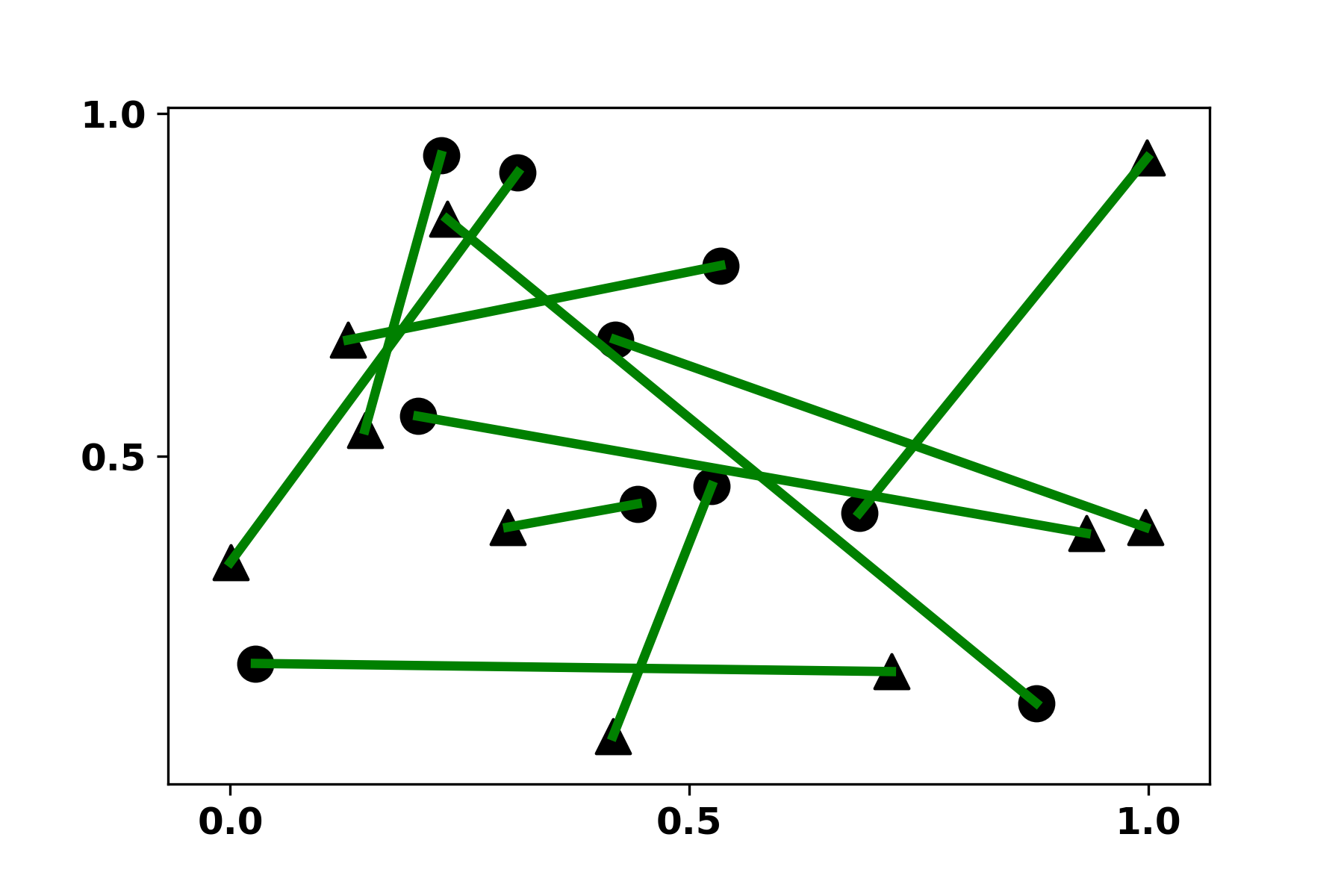}
\caption{}
\end{subfigure}%
~
\begin{subfigure}[t]{.4\textwidth}
\includegraphics[scale=0.35]{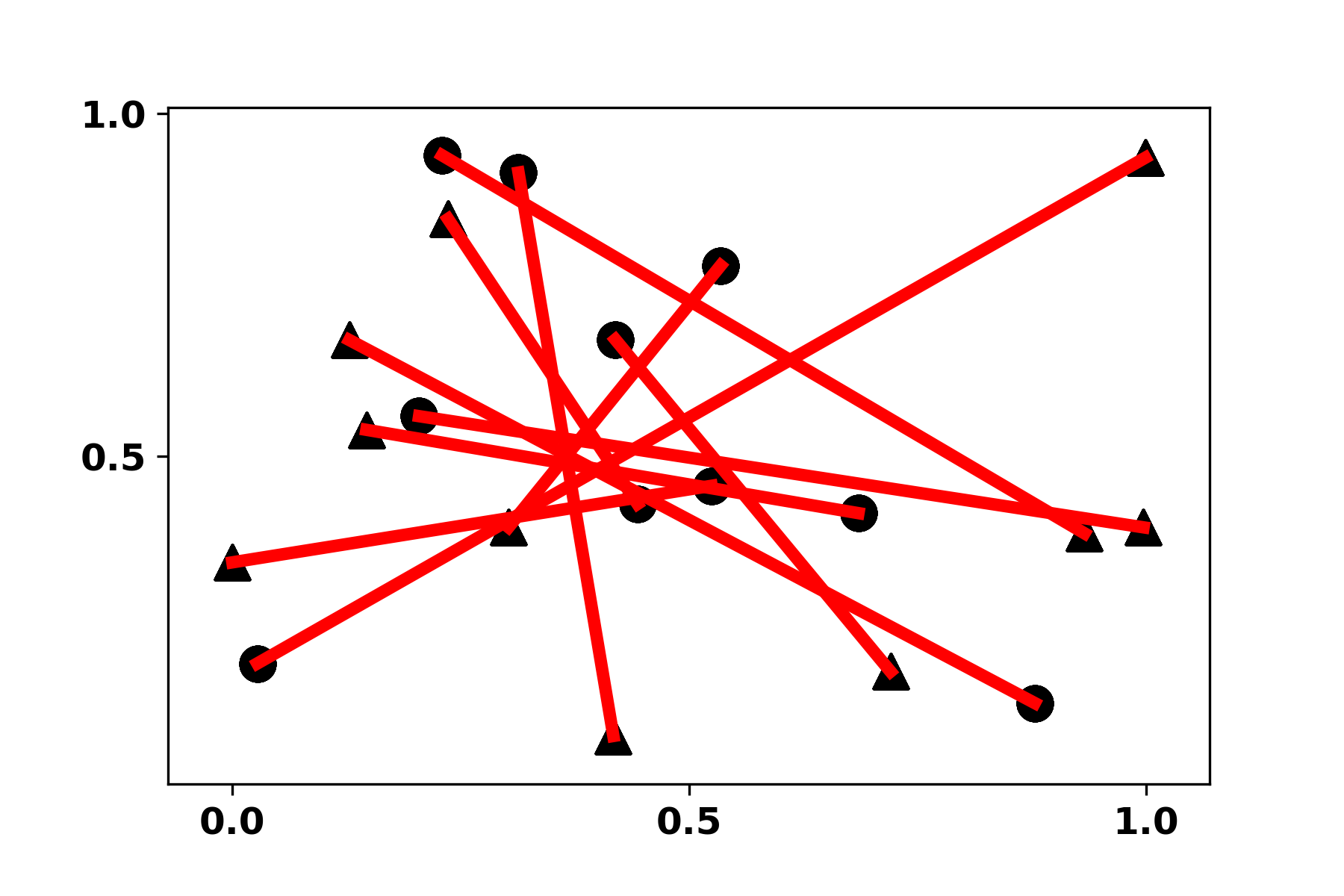}
\caption{}
\end{subfigure}
\caption{(a) The optimal matching for an instance of MWM-10. (b) A matching produced by SPG where the matching weight has an optimality ratio of 0.997. However, none of the selected match pairs are found in the optimal solution. \label{fig:app-high-ratio}}
\end{figure}

We observed that MWM becomes increasingly difficult to solve as $N$ increases and the points get more densely crowded; basically, the differences in the matching weights between candidate matchings get arbitrarily small. Additionally, SPG can find a matching that is within 0.003\% of the optimal weight but use none of the correct pairs from the optimal solution (Figure \ref{fig:app-high-ratio}). This suggests that reward shaping in the form of adding auxiliary objectives may be necessary to help solve these issues of identifiability for certain tasks.

\end{document}